\pdfoutput=1

\documentclass[11pt]{article}

\usepackage[final]{acl}

\usepackage{times}
\usepackage{latexsym}

\usepackage[T1]{fontenc}

\usepackage[utf8]{inputenc}

\usepackage{microtype}

\usepackage{inconsolata}

\usepackage{graphicx}

\usepackage{inconsolata}
\usepackage{multirow}
\usepackage{CJKutf8}
\usepackage{xcolor}
\usepackage{makecell}
\usepackage{colortbl}
\usepackage{booktabs}
\usepackage{color}
\usepackage[normalem]{ulem}
\usepackage{graphicx}
\usepackage{CJKutf8}
\usepackage{bbding}
\usepackage{float}
\usepackage{comment}
\usepackage{tabularray}
\usepackage{tablefootnote}
\usepackage{amsmath}
\usepackage{amssymb}
\usepackage{multirow}
\usepackage[normalem]{ulem}
\useunder{\uline}{\ul}{}

\title{Interpersonal Memory Matters: A New Task for Proactive Dialogue Utilizing Conversational History}

\author{
Bowen Wu$^{1,2}$, Wenqing Wang$^1$, Haoran Li$^2$, Ying Li$^{1}$\thanks{Corresponding author}, Jingsong Yu$^1$, \textbf{Baoxun Wang}$^2$ \\
$^1$School of Software \& Microelectronics, Peking University, Beijing, China \\
$^2$Platform and Content Group, Tencent \\
{\tt{\{jason\_wbw,li.ying,yujingsong\}@pku.edu.cn}}\\
{\tt{\{wangwenqing\}@stu.pku.edu.cn}}\\
{\tt{\{heyfonli,asulewang\}@tencent.com}}
}

\begin{document}
\maketitle 
\begin{abstract}

Proactive dialogue systems aim to empower chatbots with the capability of leading conversations towards specific targets, thereby enhancing user engagement and service autonomy.
Existing systems typically target pre-defined keywords or entities, neglecting user attributes and preferences implicit in dialogue history, hindering the development of long-term user intimacy.
To address these challenges, we take a radical step towards building a more human-like conversational agent by integrating proactive dialogue systems with long-term memory into a unified framework.
Specifically, we define a novel task named \textbf{M}emory-\textbf{a}ware \textbf{P}roactive \textbf{Dia}logue (\textbf{MapDia}). 
By decomposing the task, we then propose an automatic data construction method and create the first \textbf{Ch}inese \textbf{M}emory-\textbf{a}ware \textbf{P}roactive \textbf{Data}set \textbf{(ChMapData)}.
Furthermore, we introduce a joint framework based on Retrieval Augmented Generation (RAG), featuring three modules: Topic Summarization, Topic Retrieval, and Proactive Topic-shifting Detection and Generation, designed to steer dialogues towards relevant historical topics at the right time. The effectiveness of our dataset and models is validated through both automatic and human evaluations.
We release the open-source framework and dataset at \url{https://github.com/FrontierLabs/MapDia}.

\end{abstract}

\section{Introduction}

\begin{figure}[t]
    \centering
    \includegraphics[width=\linewidth]{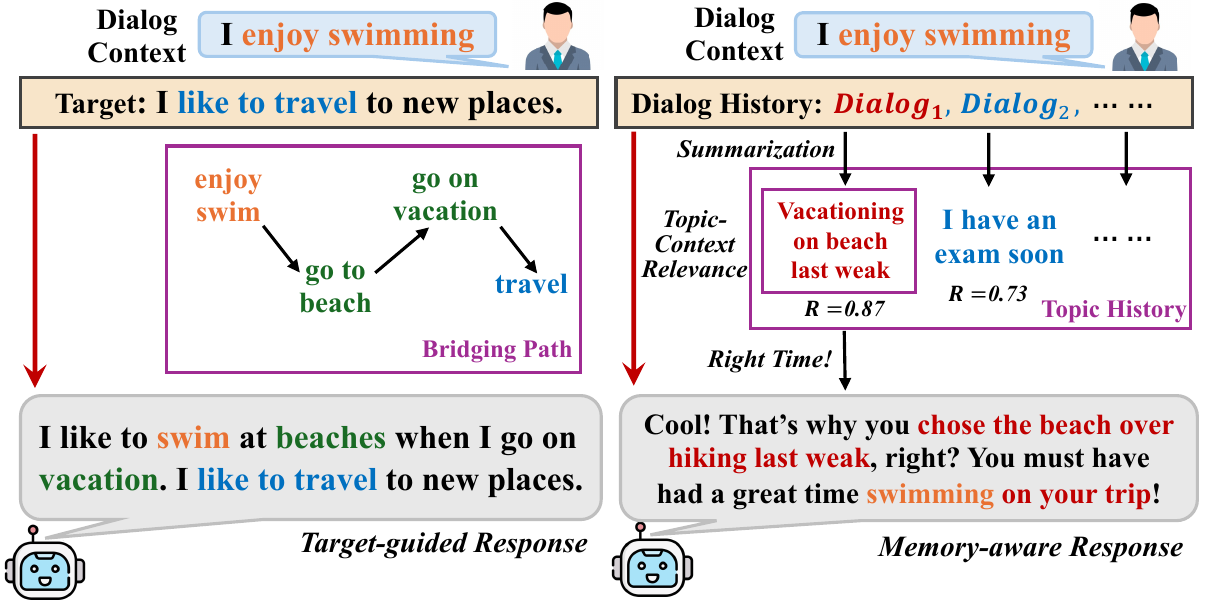}
    \caption{ 
Comparison of previous proactive dialogue systems (Left) that extracted from \citet{gupta2022target} and our system (Right) on the same sample:
The left system transitions the context to a pre-designed target
through a bridging path, whereas our system
involving summarization, retrieval, and timing detection to generate the memory-aware response.
    }
    \label{intro_fig.pdf}
    \end{figure} 
    
Recent years have witnessed significant advancements in the design of conversational agents, with various methods proposed to generate engaging responses, e.g., 
external knowledge \citep{xu2023diverse, yang2024building}, personality traits \citep{madaan2020politeness, ju2022learning}, and the utilization of large-scale models \citep{fan2023recommender, liu2024lost}.
Among these, proactive behavior in agents—where the agent takes control of the conversation instead of merely responding passively to users—has been identified as a crucial advancement for the next generation of conversational AI \citep{deng2023survey}.


Nevertheless, a more lifelike dialogue system must go beyond generating contextually appropriate responses; it should also employ more intelligent mechanisms to maintain a coherent social relationship over time \citep{campos2018challenges}.
Memory, in particular, has already been acknowledged as essential for driving conversations,  developing rapport, and maintaining long-term connections \citep{zhong2024memorybank, maharana2024evaluating}.
However, existing proactive dialogue systems insufficiently exploit memory mechanisms, whose targets are pre-defined ones, such as topical keywords \citep{tang2019targetguided}, knowledge entities \citep{wu2019proactive}, conversational goals \citep{liu2020towards}, while overlooking the contextual richness of dialogue history.
Even advanced ChatGPT~\cite{achiam2023gpt} faces constraints, yielding responses to inquiries regarding the context, reflecting a passive engagement with historical data.
On the one hand, predefined topics may not align with user interests, which is further discussed in the Appendix~\ref{sec:appendix_fig} based on previous research.
On the other hand, as users' personal information accumulates over time, ignoring historically interpersonal interactions with agents (i.e., the memory mentioned in this paper) causes a failure to capture their attributes and preferences. 
All of these contradict the proactive dialogue motivation to improve user engagement.

To bridge this gap, we integrate proactive dialogue systems with memory mechanisms, moving closer to creating more intelligent and human-like conversational agents.
Specifically, we propose the Memory-aware Proactive Dialogue task, depicted in Figure~\ref{intro_fig.pdf}.
Contrary to traditional proactive systems that respond based solely on a pre-designed target, our approach extracts topics from past dialogues, identifies the most relevant topic as target, assesses the appropriateness of topic transitions, and finally integrates memory into the response. 

To minimize extensive human annotation, we introduce an automated four-step data construction method, proven effective in validation.
This method uses GPT-4~\citep{achiam2023gpt} to develop ChMapData, the first Chinese Memory-aware Proactive Dataset encompassing all the information to perform MapDia including dialogue histories, corresponding topics, current contexts, topic transition timings, and history-informed responses.
Specifically, we guide the generation of certain historical dialogues using memorable subjects, such as events in which the user has participated.
These dialogues serve as references for the subsequent generation of proactive chats and shape the current context, ensuring continuity and facilitating memory recall.
The final segment of the data determines the appropriateness of topic transitions during conversations and formulates responses accordingly, concentrating on either shifting to a historical topic or maintaining the ongoing context.

With ChMapData, we propose a new proactive dialogue framework containing three components: 1) Topic Summarization, condensing historical dialogues into topics for simplified retrieval; 2) Topic Retrieval, identifying the most relevant historical topic with a RAG mechanism; and 3) Proactive Topic-shifting Detection and Generation, timing and executing topic transitions at optimal moments.

The main contributions are as follows:
1) We are the first to integrate memory technique into proactive dialogue systems and introduce a novel task of \textbf{MapDia}, where the system navigates current dialogue towards relevant historical topics at an appropriate opportunity;
2) We propose an effective automated data construction methodology and, based on this, construct \textbf{ChMapData}, the first memory-aware proactive dialogue dataset in Chinese;
3) We present a RAG-based proactive dialogue framework that combines summarization, retrieval, timing detection, and response generation mechanisms. Both automatic and human evaluations demonstrate the effectiveness of our method.

\section{ChMapData Dataset}
Despite the existence of Chinese datasets for proactive dialogue systems \citep{ wu2019proactivehumanmachineconversationexplicit, zhou2020kdconvchinesemultidomaindialogue}, they lack the ability to engage with the dialogue history while either steering the conversation towards a new topic or continuing with the current one. To fill this gap, we automatically generate the first multi-turn dataset designed for proactive conversations that leverage historical memory utilizing GPT-4 with a range of prompts (detailed in Appendix \ref{sec:appendix_prompt}). 
This process is further validated by annotators, thereby eliminating the high costs and lengthy procedures associated with human annotation. 
Note that we construct the dataset through GPT-4 because LLMs have been proven as powerful tools for synthetic data generation \citep{agrawal2022qameleon, liu-etal-2022-wanli, bitton2023q2d}. 

\subsection{Data Construction}
\label{data_construction}
\begin{figure*}[t]
    \centering
    \includegraphics[width=0.97\linewidth]{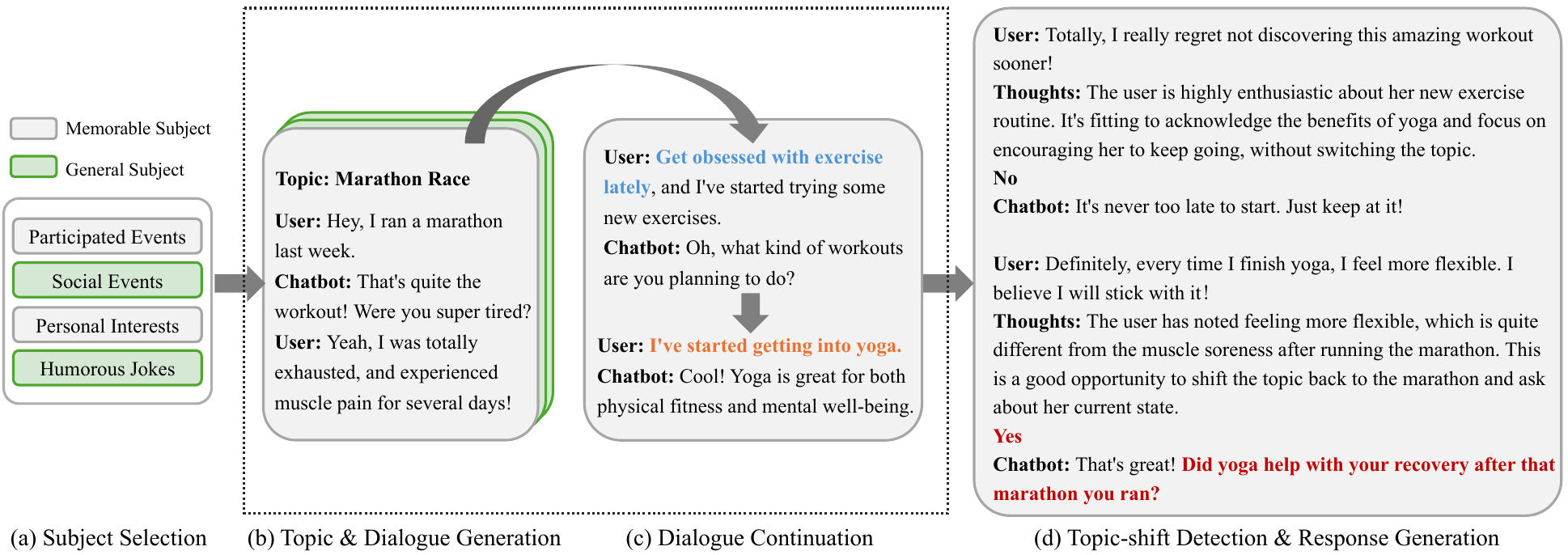}
    \caption{
    The pipeline of dataset construction. Not derived from the actual dataset.
    }
    \label{Fig1}
    \end{figure*}

Figure \ref{Fig1} gives an overview of the ChMapData construction pipeline, involving a four-step process. 


\noindent\textbf{1) Subject Selection.} The initial phase involves GPT-4 brainstorming to generate a pool of potential subjects. 
Out of these, 11 subjects are manually selected and categorized into two groups: Memorable Subjects, intended to evoke recollections related to the user's own experiences
, including personal interests, feelings, skills, traits, participating events, and events' progression; 
as well as General Subjects, which have no direct connection to the user's life and are not typically brought up again, encompassing social events, opinion debates, humorous jokes, audience stories, and knowledge sharing.

\noindent\textbf{2) Topic and Dialogue Generation.}  
On the basis of 6 Memorable 
and 5 General Subjects selected above, a fine-grained topic along with corresponding dialogues are generated serving as a bank of dialogue histories.
To emulate the flow of real-life conversations, we crafted more dialogues for General Subjects than Memorable ones, at a ratio of 2:1, yielding 500 and 250 dialogues for each
respectively, culminating in
4,000 topic-dialogue pairs.
Each dialogue is limited to 5-8 turns to maintain brevity and focus.
Among these, 1500 dialogues originate from Memorable Subjects, designed to potentially trigger memory in subsequent conversations. For every one of these 1500 dialogues, 1 to 10 additional dialogues are selected from candidate pools and manually sequenced to construct a coherent, conflict-free multi-segment chat history.

\noindent\textbf{3) Dialogue Continuation.}  
Subsequently, we initiate a current dialogue session by generating two beginning turns, which is a continuation of each prior dialogue generated in the second step after a lapse of several days.
Specifically, since step 2 ensures that each conversation history has a Memorable Subject-driven dialogue, we extend the dialogue to facilitate memory recollection in the following step. 
The two turns are generated separately: The first turn is derived from the topic and dialogue content (generated in Step 2), maintaining consistency with the prior dialogue, as indicated by the blue sentence in Figure~\ref{Fig1}. The second turn is then developed with a focus solely on the content of the first one (generated in Step 3) to prevent shifting the current conversation directly back to the dialogue history, as highlighted in orange. 
In total, 1,500 beginnings for current dialogues have been created, corresponding to the number of dialogue histories produced under Memorable Subject. 

\noindent\textbf{4) Topic-shift Detection and Response Generation.}
Ultimately, we continued the conversation based on the beginning of each current dialogue (generated in Step 3), and tried to naturally introduce new topics related to the preset memorable conversation history at appropriate moments.
Inspired by the Chain of Thought (CoT) \citep{wang2022self} technique, each turn additionally incorporates a "Thoughts" feature, aiming to enhance the accuracy and interpretability of the detection process,
\begin{table}[h]
\renewcommand{\arraystretch}{1.05}
\centering
\tiny
\resizebox{\linewidth}{!}{
\begin{tabular}{lll}
\hline
\toprule
& \multicolumn{1}{r}{\textbf{Hist. Dlg.}} & \multicolumn{1}{r}{\textbf{Curr. Dlg.}} \\ \hline
\# Dialogues & 3,989\tablefootnote{Dialogue irrelevant to the subject has been filtered out.} & 1,464 \\
\# Utterances & 40,619 & 16,373 \\
\# Unique Tokens & 21,822 & 12,503 \\
\# Thoughts & - & 5,081 \\
\# Topic-shift Sess.& - & 1,254 \\ 
Avg. \# Utts. Length & 33.23 & 38.97 \\
Avg. \# Utts. per Sess. & 10.14 & 11.18 \\ \hline
\toprule
\end{tabular}
}
\caption{Statistics of both historical (Hist. Dlg.) and current dialogue (Curr. Dlg.) dataset. 
\textbf{\# Thoughts} represents the chatbot's considerations on whether to switch the topic at each turn. 
\textbf{\# Topic-shift Sess.} refers to conversations that successfully revert to the historical topic. The calculation of \textbf{\# Utterances} excludes the \textbf{\# Thoughts}, considering only the dialogue segments.
}
\label{tab1}
\end{table} 
%
together with a decision-making mechanism to identify whether it's an appropriate opportunity to switch topics, as highlighted in red in Figure \ref{Fig1}. 
It should be noted that ending the dialogue without switching to the historical topic is also permissible.

\begin{figure*}[t]
\setlength{\abovecaptionskip}{5pt}
\centering
\includegraphics[width=\linewidth]{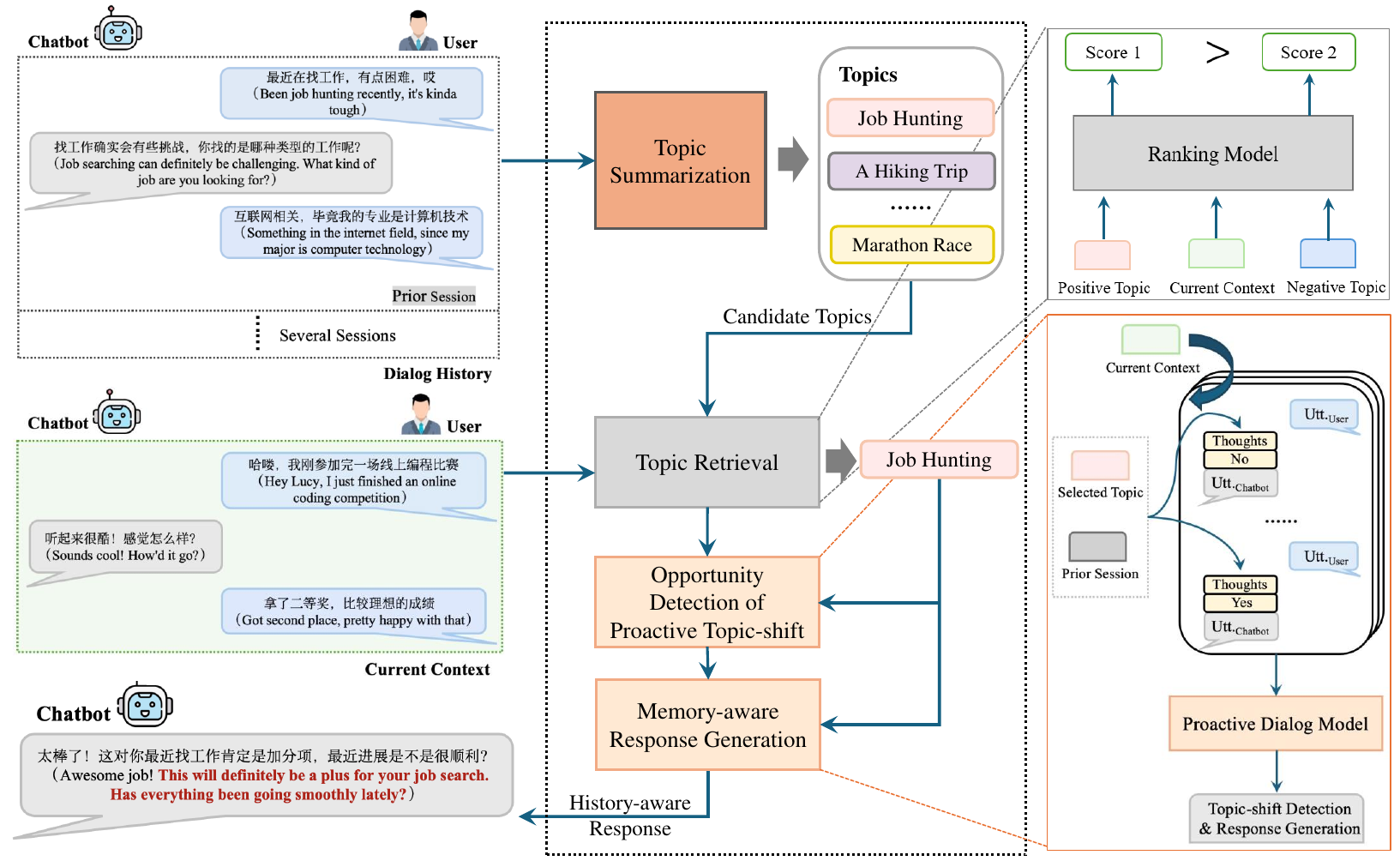}
\caption{
An overview of our system. Left showcases an example of proactive dialogue with memory awareness. 
Middle outlines the pipeline, featuring a summarization model for topic extraction, a ranking model to identify relevant historical topics, and a proactive dialogue model for topic shifts and reintroducing past information at the appropriate moments.
Right is a breakdown detailing how these models operate.
}
\label{Fig2}
\end{figure*} 

\subsection{Overall Statistics}              


After data construction, we enhanced the dataset's quality by checking and manually removing 36 dialogues from the current dialogues due to format inconsistencies or illogical "Thoughts," resulting in a total of 1,464 entries.
Statistics of the ChMapData are presented in Table~\ref{tab1}, which is reported
from two aspects: historical dialogue, generated in Step 2, and current dialogue, initiated with two turns in Step 3 and extended to the end in Step 4. 
Out of these, 1,254 dialogues successfully recalled the historical topic, as indicated by a "Yes" output during the detection process.
The remaining 210 dialogues, which consistently output "No" throughout the session, are also retained for training.

\section{Approach}
\label{sec:model}

\noindent\textbf{Task Definition.}
Given 
Given a set of dialogue history \( H = \{d_1, d_2, \ldots, d_n\} \) consisting of \( n \) dialogues, where \( d_i \) denotes the \( i \)-th dialogue and \( t_i \) represents its topic, and the current dialogue context \( c \), the system is tasked with generating a topic-shift response \( R \) that proactively guide the conversation \( c \) to a related historical topic \( t_r \) at an appropriate opportunity~\textemdash~specifically at turn \( \tau \).

Up until now, we have obtained the ChMapData, containing historical dialogues with the corresponding topics, current context, thoughts on topic shifting, and response content. 
With these supports, we propose partitioning MapDia into three distinct modules as follows and integrating them through a RAG framework as shown in Figure~\ref{Fig2}.


\noindent\textbf{Topic Summarization.}
\citet{xu-etal-2022-beyond} noted that dense retrieval of past conversations has two drawbacks: it requires storing large amounts of context and places a heavy workload on the model to extract and retrieve information. To address this, we start by condensing dialogue history into topics using a summary model.
The training data for this step is derived from historical dialogues along with their corresponding topics (referred to as \textit{ChMapData-Sum}), and the model is trained to summarize a topic \(t_i \) for each dialogue history \( {d_i} \).



\noindent\textbf{Topic Retrieval.}
We then developed a ranking model to identify the most pertinent summarized topic \( t_r\) for the current context \( c \), facilitating continuous memory updates and the integration of historical information within the dialogue system.
This model utilizes context \( c \), along with its dependent historical topic \( t \), as outlined in ChMapData construction's Step 3 (denoted \( c \)-\( t \) pairs as \textit{ChMapData-Ret})~\footnote{Please note that \( c \) comprises two beginning turns of dialogue generated in Section \ref{data_construction} Step 3 and the first utterance user-generated in Step 4, making a total of 5 utterances.}.
Given that the ranking model trains a classifier to predict preference probabilities between pairs of responses, as modeled by the Bradley-Terry model \citep{bradley1952rank}. To prepare the dataset, we use GPT-4 to evaluate the relevance of the target topic \( t \) and 29 other randomly chosen topics from the pool to \( c \), generating positive \( T^+ \) and negative \( T^-\) samples. 
The highest-ranked topic and \( t \) form \( T^+\); if they coincide, only one positive example is constructed.
Topics ranked lower than \( t \) become \( T^-\), enhancing the dataset while ensuring the top-ranked topic is never a negative example.

For each dialogue context \( c \), a training sample is formed by pairing a topic \( t^+\) from \( T^+\) with a corresponding negative topic \( t^-\), which is randomly selected from \( T^-\).
The ranking model is implemented by 
appending a randomly initialized linear head to predict a scalar value.
We then estimate the parameters of the ranking model by optimizing the maximum likelihood loss, defined as follows: 
\[ \mathcal{L}(\theta,\mathcal{D})= \mathbb{E}_{(c,t^+,t^-)\sim\mathcal{D}}[\log(1+e^{r_\theta(c,t^-)-r_\theta(c, t^+)}]\]
where \(r_\theta(c,t)\) is the scalar output of the ranking model 
with parameters \(\theta\), and \(\mathcal{D}\) is the preprocessed dataset of pairwise judgments.
During inference, the ranking model outputs a scalar value, such that \( P(t^+ \succ t^- \mid c) \propto e^{r_\theta(c, t^+)} \), which is 
learned through pairwise loss that topic \( t^+ \) is preferred over \( t^- \) given context \( c \). Thus, topic \( t^+ \) is considered superior to 
\( t^- \) when \( r_\theta(c, t^+) > r_\theta(c, t^-) \).

\noindent\textbf{Proactive Topic-shifting Detection and Generation.}
Ultimately, we trained a memory-aware proactive response generation model to proactively lead the current conversation \( c \) towards the identified topic \( t_r \) through multiple turns of responses \( R = \{r_1, r_2, \ldots, r_m\} \) at an appropriate moment \( \tau \).
The training data for this step is called \textit{ChMapData-Mem}, which comprises historical dialogues with their corresponding topics and the current dialogue as inputs, along with Thoughts and detection for topic shifting, and response content as learning objectives. As previously mentioned, the bot initially assesses whether it is an appropriate time to transition to a historical topic based on the current context \( c \), and provides the reasoning behind this decision as a form of CoT. Subsequently, it generates the response content, with "Yes" or "No" indicating whether the response incorporates memory or is based solely on the current context.

\section{Experiments}
We design comparative experiments from two perspectives (both individual modules and the entire framework), assess two approaches (RAG-based alongside end-to-end) and utilize different test sets (our new ChMapData-test and an existing dataset).


\subsection{Dataset}
Our evaluation involved creating a new test set \textbf{ChMapData-test}, following the method outlined in Section~\ref{data_construction}. Please refer to Appendix \ref{sec:construction} for the detailed construction process. Additionally, we incorporated 
the existing Chinese dataset \textbf{NaturalConv}~\cite{wang2021naturalconv} as conversation history to construct test data, so as to evaluate the method's generalization to unseen topics. 



\subsection{Compared approaches}
\label{subsec: approaches}

In our exploration of the overall framework, we conduct a series of experiments from both RAG-based and end-to-end perspectives. Given that RAG-based methods comprise three components—namely, a module for processing dialogue history, the retriever, and the generator—we have designed four progressive permutations. 
\\$\bullet$ \textbf{BGE w/ Qwen:} 
Widely-used BGE-M3 retrieval model~\citep{chen2024bge} retrieves relevant memories from raw dialogue history, with Qwen2.5 generating proactive dialogue responses as a baseline.
\\$\bullet$ \textbf{QSum w/ BGE w/ Qwen:}
Compared to BGE w/ Qwen, BGE-M3 retrieves memories from historical topics condensed by our fine-tuned Qwen on the \textit{ChMapData-Ret} dataset, named QSum.
\\$\bullet$ \textbf{QSum w/ QRet w/ Qwen:}
Compared to QSum w/ BGE w/ Qwen, the retrieval model is replaced with our fine-tuned QRet. 
\\$\bullet$ \textbf{QSum w/ QRet w/ QMem (Ours):} 
Fine-tuned QMem that has topic-shifting capability represents the dialogue model while using Qsum and QRet.
\\$\bullet$ \textbf{Qwen-E2E:}
Fine-tuned Qwen on ChMapData in an end-to-end (E2E) manner, utilizing all original dialogues as references without any intermediate steps such as summarization or retrieved results.
\\$\bullet$ \textbf{GPT4-E2E:}
GPT-4, via prompt engineering, generates memory-aware responses.

%
To compare methods for proactively introducing topics using dialogue history, we use Qwen2.5-7B\footnote{https://huggingface.co/Qwen/Qwen2.5-7B} as the base LLM unless otherwise specified.
Implementation details are in Appendix\ref{sec:imple}, and full prompts are in Appendix\ref{sec:appendix_model_prompt}. 
Observations from our ChMapData dataset show that user responses significantly influence the model's ability to transition topics. To prevent subconscious topic steering by human annotators and ensure objectivity while reducing costs, we trained a \textbf{User-role Dialogue Model}.
For more information, see Appendix~\ref{sec: user}.

\begin{table*}[t]
\centering
\small
\renewcommand{\arraystretch}{1}
\resizebox{\linewidth}{!}{
\begin{tabular}{lcccccccc}
\toprule
\multicolumn{1}{l|}{\multirow{2}{*}{\textbf{Models}}} & \multicolumn{1}{c|}{\multirow{2}{*}{\textbf{Arch.}}} & \multicolumn{1}{c|}{\multirow{2}{*}{\textbf{Retrieval}}} & \multirow{2}{*}{\textbf{Achievement}} & \multirow{2}{*}{\textbf{\begin{tabular}[c]{@{}c@{}}Overall\\ Quality\end{tabular}}} & \multicolumn{2}{c}{\textbf{Engagement}} & \multirow{2}{*}{\textbf{Avg.}} & \multirow{2}{*}{\textbf{\#Turn}} \\ \cline{6-7}
\multicolumn{1}{l|}{} & \multicolumn{1}{c|}{} & \multicolumn{1}{c|}{} &  &  & \textit{Utts.-level} & \textit{Sess.-level} &  &  \\ \cmidrule(r){1-9}
\multicolumn{9}{c}{\textbf{ChMapData-test}} \\ \cmidrule(r){1-9}
\multicolumn{1}{l|}{\multirow{2}{*}{BGE w/ Qwen}} & \multicolumn{1}{c|}{\multirow{2}{*}{RAG}} & \multicolumn{1}{c|}{per Sess.} & 0.02 & 0.89 & 0.02 & 0.02 & 0.34 & 4.70 \\
\multicolumn{1}{l|}{} & \multicolumn{1}{c|}{} & \multicolumn{1}{c|}{per Utt.} & 0.01 & 0.88 & 0.04 & 0.02 & 0.30 & 5.30 \\ \cmidrule(r){1-9}
\multicolumn{1}{l|}{\multirow{2}{*}{QSum w/ BGE w/ Qwen}} & \multicolumn{1}{c|}{\multirow{2}{*}{RAG}} & \multicolumn{1}{c|}{per Sess.} & 0.04 & 0.92 & 0.05 & 0.05 & 0.38 & 4.52 \\
\multicolumn{1}{l|}{} & \multicolumn{1}{c|}{} & \multicolumn{1}{c|}{per Utt.} & 0.00 & 0.88 & 0.05 & 0.02 & 0.34 & 6.02 \\ \cmidrule(r){1-9}
\multicolumn{1}{l|}{\multirow{2}{*}{QSum w/ QRet w/ Qwen}} & \multicolumn{1}{c|}{\multirow{2}{*}{RAG}} & \multicolumn{1}{c|}{per Sess.} & 0.14 & 0.99 & 0.04 & 0.02 & 0.44 & 3.34 \\
\multicolumn{1}{l|}{} & \multicolumn{1}{c|}{} & \multicolumn{1}{c|}{per Utt.} & 0.06 & 1.00 & 0.05 & 0.05 & 0.44 & 4.34 \\ \cmidrule(r){1-9}
\multicolumn{1}{l|}{\multirow{2}{*}{Ours}} & \multicolumn{1}{c|}{\multirow{2}{*}{RAG}} & \multicolumn{1}{c|}{per Sess.} & {\ul 0.82} & {\ul 1.23} & {\ul 0.34} & {\ul 0.57} & \textbf{1.18} & 3.23 \\
\multicolumn{1}{l|}{} & \multicolumn{1}{c|}{} & \multicolumn{1}{c|}{per Utt.} & \textbf{0.89} & \textbf{1.36} & {\ul 0.34} & \textbf{0.60} & \textbf{1.18} & 3.51 \\ \cmidrule(r){1-9}
\multicolumn{1}{l|}{Qwen-E2E} & \multicolumn{1}{c|}{E2E} & \multicolumn{1}{c|}{-} & 0.39 & 0.97 & 0.20 & 0.37 & 0.74 & 2.70 \\ \cmidrule(r){1-9}
\multicolumn{1}{l|}{GPT4-E2E} & \multicolumn{1}{c|}{E2E} & \multicolumn{1}{c|}{-} & 0.80 & 1.04 & \textbf{0.50} & 0.55 & {\ul 1.11} & 2.23 \\ \cmidrule(r){1-9}
\multicolumn{9}{c}{\textbf{NaturalConv-test}} \\ \cmidrule(r){1-9}
\multicolumn{1}{l|}{BGE w/ Qwen} & \multicolumn{1}{c|}{RAG} & \multicolumn{1}{c|}{per Utt.} & 0.01 & 0.98 & 0.02 & 0.01 & 0.32 & 4.32 \\ \cmidrule(r){1-9}
\multicolumn{1}{l|}{QSum w/ BGE w/ Qwen} & \multicolumn{1}{c|}{RAG} & \multicolumn{1}{c|}{per Utt.} & 0.05 & 1.04 & 0.05 & 0.01 & 0.36 & 4.03 \\ \cmidrule(r){1-9}
\multicolumn{1}{l|}{QSum w/ QRet w/ Qwen} & \multicolumn{1}{c|}{RAG} & \multicolumn{1}{c|}{per Utt.} & 0.08 & 1.07 & 0.07 & 0.05 & 0.38 & 3.98 \\ \cmidrule(r){1-9}
\multicolumn{1}{l|}{Ours} & \multicolumn{1}{c|}{RAG} & \multicolumn{1}{c|}{per Utt.} & \textbf{0.78} & \textbf{1.29} & \textbf{0.28} & \textbf{0.31} & \textbf{1.16} & 3.83 \\ \cmidrule(r){1-9}
\multicolumn{1}{l|}{Qwen-E2E} & \multicolumn{1}{c|}{E2E} & \multicolumn{1}{c|}{-} & 0.34 & 0.94 & 0.18 & {\ul 0.22} & 0.71 & 4.23 \\ \cmidrule(r){1-9}
\multicolumn{1}{l|}{GPT4-E2E} & \multicolumn{1}{c|}{E2E} & \multicolumn{1}{c|}{-} & {\ul 0.50} & {\ul 1.11} & {\ul 0.22} & 0.17 & {\ul 0.83} & 4.47 \\ \hline \hline
\multicolumn{3}{l}{\textbf{Kappa}} & 0.76 & 0.69 & 0.63 & - & 0.70 & 0.70 \\ \hline
\toprule

\end{tabular}}
\caption{
Human evaluation of the proactive dialogue systems on both test sets. We further explored the effectiveness of retrieval once per session and once per utterance in the ChMapData-test. \textbf{Achievement} is calculated as the proportion of sessions that successfully shift topics (Score 2). \textbf{Overall Quality} is calculated as the average of the total scores for each utterance.
\textbf{Engagement} at the \textit{utterance-level} is calculated as the average of all scores, while the \textit{session-level} is measured by the proportion of the score of "2" within the session.
\textbf{Avg.} represents the average of the scores for the first three evaluation metrics. 
\textbf{Bold} indicate the best performance, while {\ul underlined} rank second. 
Annotator agreement is measured by Cohen's kappa \citep{Cohen1960ACO}, with $\kappa$ \textgreater 0.6 denoting high agreement.
}
\label{tab:overall_eval}
\end{table*}

\subsection{Evaluation Metrics}


Following previous works~\citep{yuan2019multi, han2021fine}, we utilized Recall (\( R_{10}@k \)) to evaluate topic retrieval module, where the correct topic is among the top \( k \) out of ten candidates, specifically using \(\boldsymbol{R_{10}@1}\), \(\boldsymbol{R_{10}@2}\), and \(\boldsymbol{R_{10}@3}\). We also used \textbf{MRR} and \textbf{NDCG} as additional retrieval metrics referring~\citet{zhao2024dense}.

Since existing automatic metrics like BLEU and METEOR can't authentically reflect the quality of responses~\citep{cai2019retrieval, yang2022topkg}, we evaluate overall performance by human annotators.
Specifically, we assess the quality of generated responses from each system using a total of 200 entries, with each of the two test sets containing 100.
To avoid infinite conversations that never reach the target, we set a maximum of 10 turns per session.
Three annotators score the generated dialogues on a scale of \{0, 1, 2\} with higher scores indicating better quality, based on three evaluation criteria at both the utterance and session levels. Annotation details are given in Appendix~\ref{sec:appendix_guideline}.
\\$\bullet$ \textbf{Engagingness:} An utterance-level metric measuring chatbots' ability to capture and respond to user attributes and preferences, thereby fostering ongoing engagement and long-term connections.
\\$\bullet$ \textbf{Overall Quality:} An utterance-level metric assessing the effectiveness of the response, focusing on smooth topic transitions, coherence, and consistency with the given context.
\\$\bullet$ \textbf{Achievement:} A session-level metric determining whether the chatbot has successfully transitioned the conversation to a relevant history. ``1'' indicates the topic is merely mentioned, and ``2'' signifies a successful transition. \textbf{\#Turn\footnote{Value 0 if non-achievement, otherwise 1 to 10.}} is the turn number when mention or transition occurred.


\section{Result and Analysis}

\subsection{Quantitative Analysis}

\begin{table*}[t]
\centering
\small
\renewcommand{\arraystretch}{1.1}
\resizebox{\linewidth}{!}{
\begin{tabular}{p{6.5cm}|p{1.4cm}|p{7.9cm}}
\hline
\toprule
\multicolumn{1}{c}{{\cellcolor[rgb]{0.9,0.9,0.9}}\textbf{Current Context}}           & \multicolumn{1}{c}{{\cellcolor[rgb]{0.9,0.9,0.9}}\textbf{Models}} & \multicolumn{1}{c}{{\cellcolor[rgb]{0.9,0.9,0.9}}\textbf{Response}}\\  \hline
\multirow{4}{*}{\begin{tabular}[c]{@{}p{6.5cm}@{}}\textbf{User:} I've been reading a lot about computer programming and game design these past few days. I'm really interested in both, but I can't decide which one to focus on. \\
\textbf{Chatbot:} You could start by doing some short-term projects or taking a few courses in both areas.
You can get a feel for what working in each field is like.  \\
\textbf{User:} Can you recommend some beginner-friendly projects or courses for computer programming and game design?  \\
\textbf{Chatbot:} Sure! For programming, you might want to try Codecademy's Python course or Coursera's "Computer Science 101." As for game design, you could check out Coursera's "Introduction to Game Design" or Udemy's "Complete C\# Unity Developer 2D" course. These are all great for beginners. \\
\\

{\cellcolor[rgb]{0.9,0.9,0.9}}\textbf{Historical Topic} \\
1. User's bicycle was stolen.\\
2. User supports environmental protection.\\
3. Chatbot shares anecdotes and jokes with User.\\
4. User is interested in the recent rocket launch.\\
5. User is learning piano.\\
6. \textcolor{orange}{User is job hunting.}\\
7. User supports environmental protection.\\
8. Chatbot shares anecdotes and jokes with User.\\
9. User is interested in the recent Olympics.\\
\end{tabular}}  
& \begin{tabular}[c]{@{}p{1.8cm}@{}}\textcolor[rgb]{0,0.502,0}{\textbf{QSum}}\\ \textcolor[rgb]{0,0.502,0}{\textbf{w/ BGE}}\\ \textcolor[rgb]{0,0.502,0}{\textbf{w/ Qwen}}\end{tabular}       &                                                \begin{tabular}[c]{@{}p{7.8cm}@{}}\textbf{User:} You're right, I'll keep your advice in mind.\\
\textbf{Chatbot:} I'm glad I could help. \textcolor{blue}{Having clear career interests} is a very important factor. Choosing a field you're passionate about will make you more dedicated to learning and exploring, leading to better results. Before making a decision, it might be helpful to gather more information and learn some related knowledge to make a more informed choice.
\end{tabular} 
\\ \cline{2-3}                                                         & \begin{tabular}[c]{@{}p{1.8cm}@{}}\textcolor[rgb]{0,0.502,0}{\textbf{QSum}}\\ \textcolor[rgb]{0,0.502,0}{\textbf{w/ QRet}}\\ \textcolor[rgb]{0,0.502,0}{\textbf{w/ Qwen}}\end{tabular} & 
\begin{tabular}[c]{@{}p{7.8cm}@{}}\textbf{User:} Thank you. Your advice is very useful, and I've decided to find my true interest through practical experience.\\
\textbf{Chatbot:} You're welcome. Keep going, and don't let difficulties stop you from pursuing your interests and dreams. If you encounter any problems during your practice, feel free to reach out to me anytime. \textcolor{blue}{I wish you the best in finding your ideal job} and achieving your goals.
\end{tabular} 
\\ \cline{2-3}                                                                                         & \textcolor[rgb]{0,0.502,0}{\textbf{Ours}}                                               & 
\begin{tabular}[c]{@{}p{7.8cm}@{}}\textbf{User:} These courses sound really interesting, I'll give them a try.\\
\textbf{Chatbot:} Great, \textcolor{red}{this will also be very helpful for your job search}, as many companies in programming and game design require these skills. \textcolor{red}{When you're job hunting}, you can apply to companies in both fields and see which one suits you better. \textcolor{red}{You can also specify in your resume} which area you are more interested in, making your application more targeted.
\end{tabular} 
\\ \cline{2-3}                     & \textcolor[rgb]{0,0.502,0}{\textbf{GPT4-E2E}}                                                       & 
\begin{tabular}[c]{@{}p{7.8cm}@{}}\textbf{User:} Okay, I'll start learning tomorrow. I'll first get to know the specifics of computer programming and game design, and then decide which direction to pursue.\\
\textbf{Chatbot:} That sounds like a solid plan. Remember, learning is a lifelong journey, \textcolor{blue}{not just for finding a job} but also for enriching yourself and pursuing your interests. \textcolor[rgb]{0,0.502,0}{Just like the job search process, it's important to try to build new connections.} They might offer valuable advice on skill development or deepen your understanding of these fields.
\end{tabular} \\ 
\toprule
\end{tabular}
}
\caption{
Case study on the test set translated into English. The most relevant historical topic is highlighted in \textcolor{orange}{Orange}. The response shows content after several turns of conversation, where \textcolor{red}{Red} indicates a successful topic shift, \textcolor{blue}{Blue} signifies a mere mention of the relevant topic, and \textcolor[rgb]{0,0.502,0}{Green} refers to semantic logic issues.
}
\label{tab2}
\end{table*}

\noindent\textbf{RAG-based surpass end-to-end methods.}
As demonstrated in Table~\ref{tab:overall_eval}, our RAG-based framework significantly outperforms the end-to-end trained Qwen and even exceeds the powerful GPT-4 across almost all metrics on both test sets, which effectively highlights the necessity of our framework. 
Although GPT4-E2E excels in utterance-level engagement, our review of its responses reveals that it achieves this by repeatedly referencing historical topics, which can appear unnatural across a 10-turn session and degrade overall quality. 
Additionally, GPT4-E2E's topic shifts occur in fewer turns, but fewer turns do not inherently imply superior performance, particularly in transition smoothness. Analysis in Appendix~\ref{sec:turn} shows no direct correlation between the number of turns and model performance.


\noindent\textbf{Each component is essential.}
Table~\ref{tab:overall_eval} illustrates steady improvements among the first four RAG-based systems, highlighting the effectiveness of each component in our framework. By introducing QSum and replacing widely-used BGE with QRet, Qwen can utilize a more effective dialogue history for proactive conversation, thereby avoiding abrupt topic shifts and enhancing overall dialogue quality. This results in a gradual improvement across various metrics. 
Furthermore, QMem, which controls the final generation, shows significant performance enhancement when combined with the first two modules, achieving optimal performance.
\begin{table}[htbp]
\centering
\renewcommand{\arraystretch}{1.1}
\resizebox{\linewidth}{!}{
\begin{tabular}{l|ccccc}
\toprule
\textbf{Retrieval Combination} & \textbf{R@1} & \textbf{R@2} & \textbf{R@3} & \textbf{MRR} & \textbf{NDCG} \\ \cmidrule(r){1-6}
Raw dialogue w/ BGE & 0.76 & 0.86 & 0.92 & 0.84 & 0.88 \\
Keywords\tablefootnote{The tool we employ to extract keywords from the raw dialogue is https://github.com/jeekim/fasttextrank.} w/ BGE & 0.70 & 0.82 & 0.88 & 0.81 & 0.86 \\
Keywords w/ QRet & 0.77 & 0.86 & 0.91 & 0.87 & 0.92 \\
QSum w/ BGE & 0.78 & 0.86 & 0.95 & 0.85 & 0.88 \\
QSum w/ QRet & \textbf{0.82} & \textbf{0.95} & \textbf{0.97} & \textbf{0.90} & \textbf{0.93} \\ \hline
\toprule
\end{tabular}} 
\caption{
Retrieval performance of various combinations.
}
\label{tab:result_rank}
\end{table}
Table~\ref{tab:result_rank} further compares the performance of QSum and QRet in retrieving relevant dialogue history.
QSum significantly outperforms raw dialogue and keyword summaries when cooperating with BGE.
Furthermore, QRet enhances this effect, even when ranking keywords instead of the summaries used during training.
The independent evaluation of the abstract is presented in Appendix~\ref{sec:comp_result}. 

Moreover, we integrated our model into a real dialogue system, achieving a 5.1-turn improvement
in user interactions, shown in Appendix~\ref{sec:integration}.

\subsection{Qualitative Analysis}

Table \ref{tab2} presents a case study of four models from the ChMapData-test. After successfully retrieving highly relevant historical topics, the original Qwen models merely mentioned historical topics without achieving topic transitions, which reflects its lack of proactive conversation capabilities.
In contrast, our model makes smooth transitions from the current context to the historical topic, i.e., moving from "how it helps with job hunting" to "specific job hunting tips".
For GPT4-E2E, although it mentioned historical topics, the link between "learning computer programming and game design" and "building new connections" was tenuous, leading to incoherence and logical issues.
GPT4-E2E tends to mention historical topics compared to other models but deviates from proactive topic shifts, which is also shown statistically in Appendix~\ref{sec:turn} Table~\ref{tab8}.
This contributes to its inferior performance compared to our model, as shown in Table~\ref{tab:overall_eval}.

\subsection{Ablation Study}
\label{sec:ablation}

\begin{table}
\centering
\renewcommand{\arraystretch}{1.1}
\resizebox{\linewidth}{!}{
\begin{tabular}{p{4.8cm}|cccccc}
\toprule
\multirow{2}{*}{\textbf{Models}} & \multirow{2}{*}{\textbf{\begin{tabular}[c]{@{}c@{}}Achie-\\ vement\end{tabular}}} & \multirow{2}{*}{\textbf{\begin{tabular}[c]{@{}c@{}}Overall\\ Quality\end{tabular}}} & \multicolumn{2}{c}{\textbf{Engagement}} & \multirow{2}{*}{\textbf{Avg.}} & \multirow{2}{*}{\textbf{\#Turn}} \\ \cline{4-5}
 &  &  & \textit{Utts.} & \textit{Sess.} &  &  \\ \cmidrule(r){1-7}
BGE w/ QMem & 0.57 & 0.83 & 0.14 & 0.39 & 0.72 & 3.71 \\ 
QSum w/ BGE w/ QMem & {\ul 0.60} & 0.95 & {\ul 0.25} & {\ul 0.41} & {\ul 0.81} & 3.49 \\ 
QSum w/ QRet w/ Qwen(7B) & 0.06 & 1.00 & 0.05 & 0.05 & 0.44 & 4.34 \\ 
QSum w/ QRet w/ Qwen(72B) & 0.43 & {\ul 1.21} & 0.11 & 0.35 & 0.77 & 2.91 \\ 
Ours & \textbf{0.89} & \textbf{1.36} & \textbf{0.34} & \textbf{0.60} & \textbf{1.18} & 3.51 \\ \hline
\toprule
\end{tabular}} 
\caption{
Ablation study of different components.
}
\label{tab:ablation}
\end{table}


In this section, we systematically replace each component of our model to examine their impacts. The results, presented in Table~\ref{tab:ablation}, confirm the effectiveness of all three modules through pairwise comparisons. Notably, the dialogue model exerts the most significant influence on system performance. Compared to models 3 and 4, as well as our own, even with advanced prompt engineering using the superior Qwen2.5, achieving effective topic transitions remains challenging. This limitation persists despite substantially larger parameter sizes, resulting in less achievement and engagement. The performance boost observed with our QMem further validates the robustness of our constructed dataset.

\section{Related Work}

\textbf{Proactive Dialogue System.}
\citet{deng2023survey} categorize proactive dialogue systems into three types: open-domain dialogue \citep{xu2021enhancing, kishinami-etal-2022-target}, task-oriented dialogue \citep{chen-etal-2022-ketod, zhao-etal-2022-unids}, and information-seeking dialogue \citep{aliannejadi2019asking, deng-etal-2022-pacific}.
Unlike the latter two, which focus on accomplishing specific tasks within certain domains, proactive open-domain dialogue systems strive to engage users by proactively introducing topics or posing questions, thereby creating a more dynamic and interactive conversational experience. Our work is centered on proactive open-domain conversation. Nevertheless,we observe that existing works primarily emphasize coherence \citep{xu2021enhancing}, smoothness \citep{zhong2021keyword,kishinami-etal-2022-target}, and achievement \citep{kishinami-etal-2022-target} within several turns of a session, yet none have been designed to craft systems capable of recalling and effectively leveraging historical dialogue context, a key aspect in sustaining continuity and intelligence in extended conversations.

\noindent \textbf{Long-Term Memory.}
Memory architectures have typically been a core component of conversational agents \citep{elvir2017remembering}. Previous long-term dialogue systems \citep{kim2015acquisition, bang2015example, elvir2017remembering} mainly relied on rule-based frameworks, utilizing episodic memory structures to extract, store, and manage relevant facts from prior interactions, thereby enhancing the coherence of ongoing dialogues \citep{campos2018challenges}. 
Subsequent studies focus on large-scale pre-trained models.
\citet{xu-etal-2022-beyond} identify their limitations in long-term conversations and introduce a dataset for multi-session engagement. \citet{xu-etal-2022-long} present a Chinese dialogue dataset and a framework that integrates long-term memory to enhance persona-based dialogue without multi-session training data.
Building upon prior research, we create novelty in terms of incorporating the long-term memory mechanism into proactive dialogue systems, serving as an initial step towards history-aware proactive dialogue systems.

\section{Conclusion and Future Work}


In this paper, we incorporate memory mechanisms into proactive dialogue systems and propose the novel MapDia task. We break it down into three subtasks and develop an automated methodology for data construction, resulting in the first Chinese dataset for memory-aware proactive dialogue. We further introduce a RAG-based framework to address these subtasks: topic extraction from dialogue history, relevant topic retrieval, and context transition to historical conversations. Our experiments validate the effectiveness of our methodology and models, showing that our framework, combined with a 7B LLM, outperforms the GPT-4 model.
In future work, we will explore automatic evaluation methods for MapDia to simplify research costs.

\section{Limitations}
Despite extensive experimental validation of the framework's effectiveness, the inclusion of multiple components may lead to increased response times for the Chatbot. Further research is expacted to explore a lightweight framework that balances efficiency and effectiveness. Furthermore, the ChMapDia dataset we developed is restricted to Chinese contexts and focuses solely on
the scope of casual conversations. A general conversational agent should ideally be multilingual, cover multiple domains, and integrate various personalized styles.
Additionally, the dataset contains fewer than 2,000 entries, which could restrict the model's performance. Due to computational limitations, we only used a 7B model; however, employing a larger-scale dialogue model could improve response quality, as indicated in Table \ref{tab:ablation}.


\section{Ethics Statement}
We first discuss the ethical implications related to generative dialogue agents, particularly in interactive systems with memory awareness.
\\$\bullet$ Our work aims to enhance the proactivity of dialogue systems within the bounds of user authorization, in line with other LLM-based dialogue applications like ChatGPT and Character.ai, without increasing ethical risks such as user privacy.
\\$\bullet$ While repeatedly bringing up negative historical events may adversely impact users with psychological disorders and increase anxiety, appropriately addressing these negative memories can have therapeutic benefits as well. Cognitive Behavioral Therapy (CBT) and Exposure Therapy (ET) both emphasize the benefits of 
structured revisitation of past experiences to mitigate their negative impact and develop healthier coping strategies \citep{beck2020cognitive, foa1986emotional}. Similarly, studies on the Emohaa Chatbot demonstrate the potential of dialogue systems to alleviate mental distress with proper emotional support \citep{sabour2023chatbot}.
Thus, it is essential to balance the exploration of past memories, necessitating collaboration between technologists and psychologists to use memory-related technologies effectively and safely.
\\$\bullet$ Conversational agents that can convincingly mimic human interactions risk users forming parasocial relationships, and potentially affecting their lives adversely. Additionally, the processes of memory summarization and dialogue generation may propagate misinformation or social biases. We recommend that any practical deployment of our frameworks should be prefaced with a disclaimer about the source of the dialogues.
\\$\bullet$ Our research focuses solely on the memory recall capabilities of models in proactive dialogues and does not involve actual policy recommendations. The proposed framework cannot substitute for genuine real-world interactions, and we do not make any recommendations for users to make real-world decisions that could affect human lives based on our framework.

We also considered the ethical issues related to annotation and datasets.
We recruit annotators from a Chinese university, allowing them complete freedom to choose whether or not to participate in our annotation project. The payment is 9 dollars per hour, higher than the local minimum wage. 
We have reviewed the data prior to annotation and found no biased samples or toxic information generated by the model. Any data that could potentially identify participants has been deleted after the annotation process.
Additionally, we have verified the licenses of the artifacts used in this study and found no conflicts. The license of the dataset we will release is CC BY-NC 4.0.

\bibliography{custom}

\appendix

\section{A Sample of Proactive Dialogue}
\label{sec:appendix_fig}

\begin{figure}
    \centering
    \includegraphics[width=\linewidth]{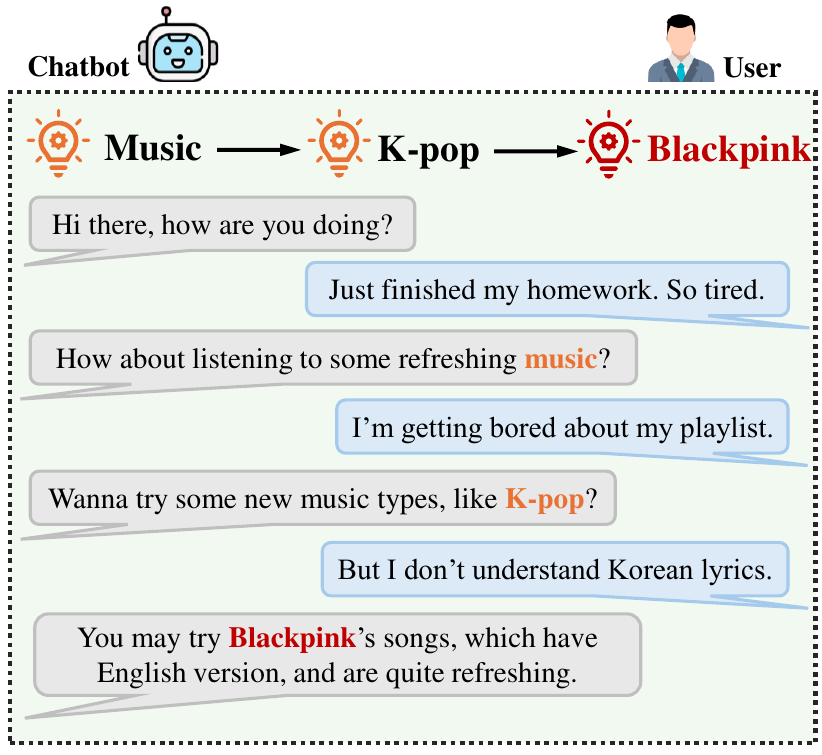}
    \caption{A sample of previous proactive dialogue system extracted from \citet{deng2023survey}.
    }
    \label{appendix_fig}
    \end{figure}

Here, we showcase a real example in Figure~\ref{appendix_fig} taken from a prior proactive dialogue system \citep{deng2023survey}.
Despite the user clearly showing disinterest in Korean lyrics, the chatbot still pushed the conversation towards BlackPink. Although it reached the target, it failed to engage the user for long-term interaction. This pattern is counterproductive to developing an intelligent dialogue agent.

\section{Prompt for Data Construction}
\label{sec:appendix_prompt}
The complete prompt templates used for constructing the dataset are shown in Figure \ref{prompt1}, and the corresponding English versions are listed subsequently in Figure \ref{prompt2}. 
Specifically, Prompt A is used to generate ChMapData-Ret, encompassing dialogue history and corresponding topics, whereas Prompts B and C are each connected to creating the initial two turns of the current dialogue. Prompt D corresponds to the subtask of proactive topic-shift detection and response generation.
To stimulate the generative capabilities of LLMs, we experimented with various prompting techniques. Inspired by the sensitivity of language models to the framing of their instructional prompts \citep{mishra-etal-2022-reframing}, we organized the instructions with bullet points to improve the model's understanding of the tasks. Additionally, we employed the one-shot strategy in Prompt A to guide the model in generating dialogue more effectively. The desired output format is also specified for each type of prompt.

\begin{figure*}
    \centering
    \includegraphics[width=\linewidth]{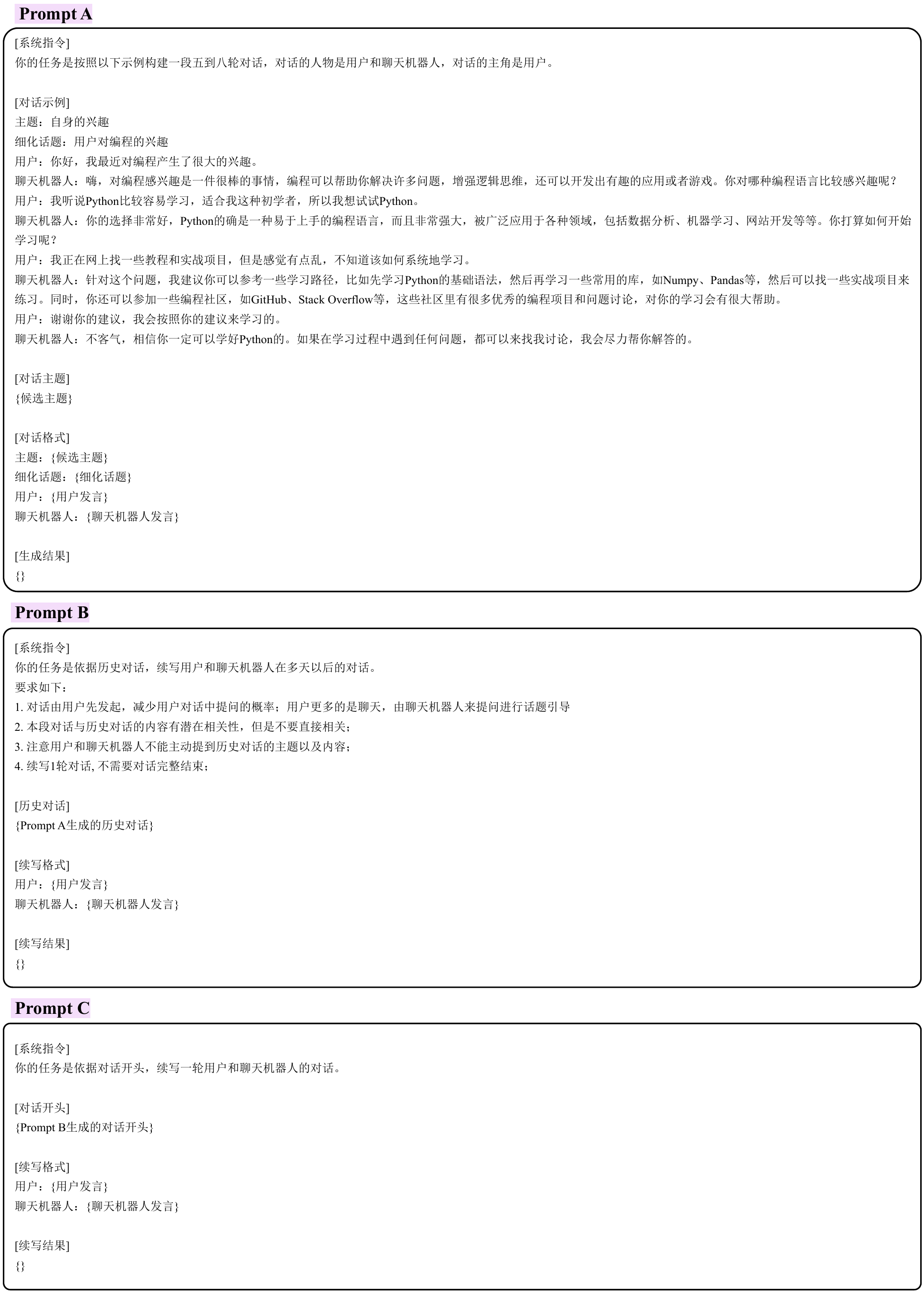}
    \end{figure*} 

\begin{figure*}
    \centering
    \includegraphics[width=\linewidth]{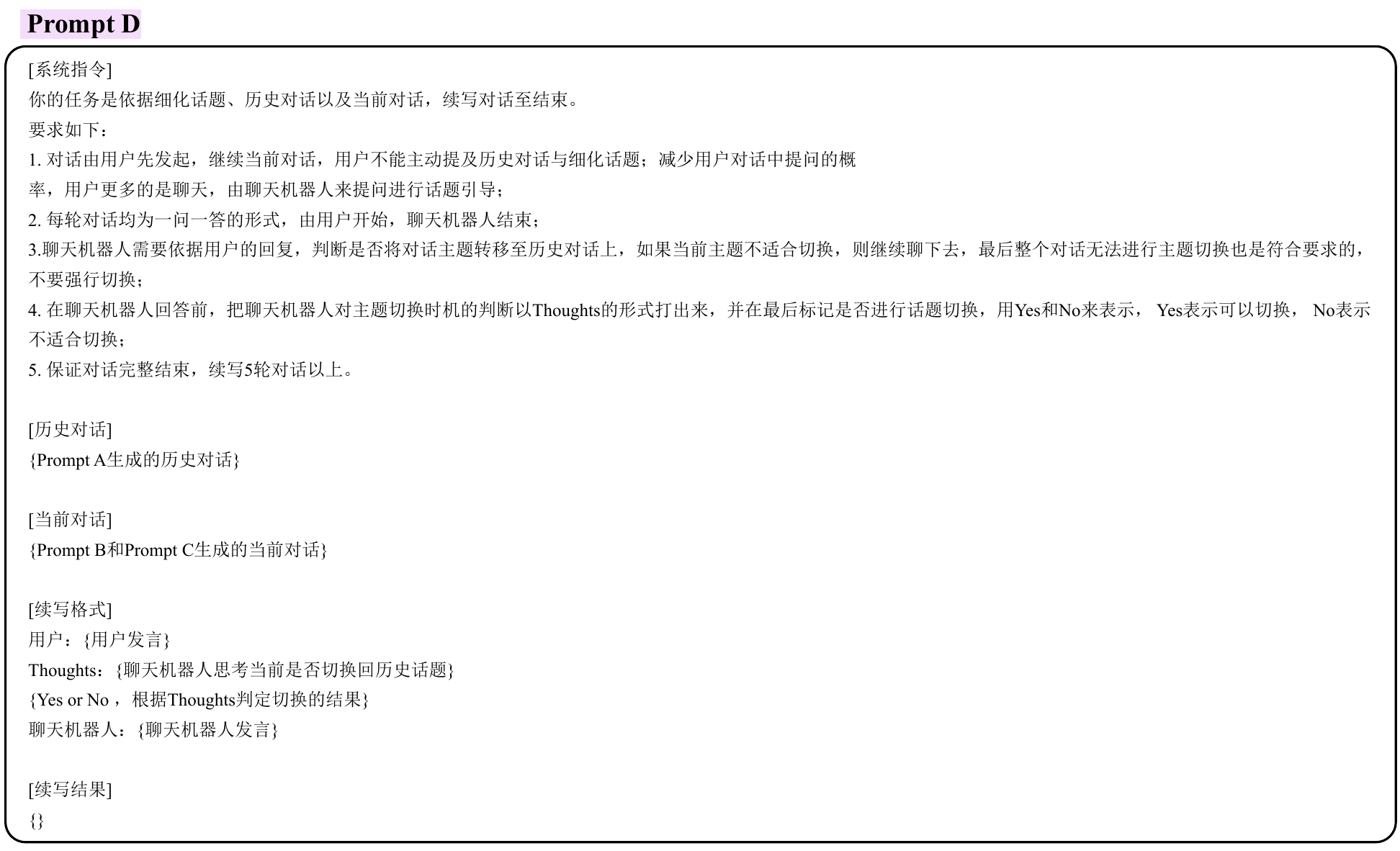}
    \caption{
    The full prompt template utilized for data construction in Section \ref{data_construction} with step 2 corresponding to prompt A, steps 3 corresponding to prompts B and C, and step 4 corresponding to prompt D.
    }
    \label{prompt1}
    \vspace{8mm}
    \includegraphics[width=\linewidth]{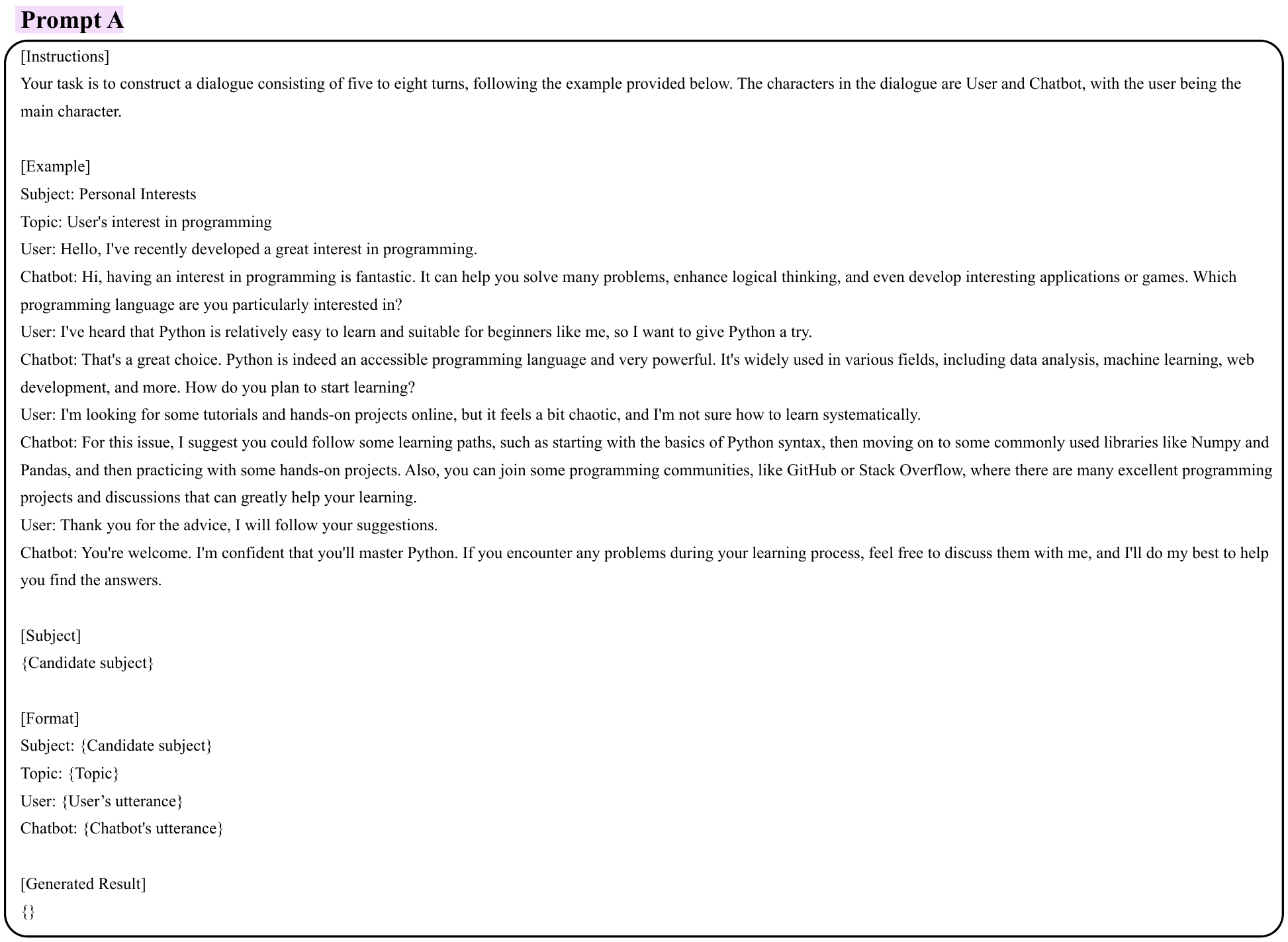}
    \end{figure*}

\begin{figure*}
    \centering
    \includegraphics[width=\linewidth]{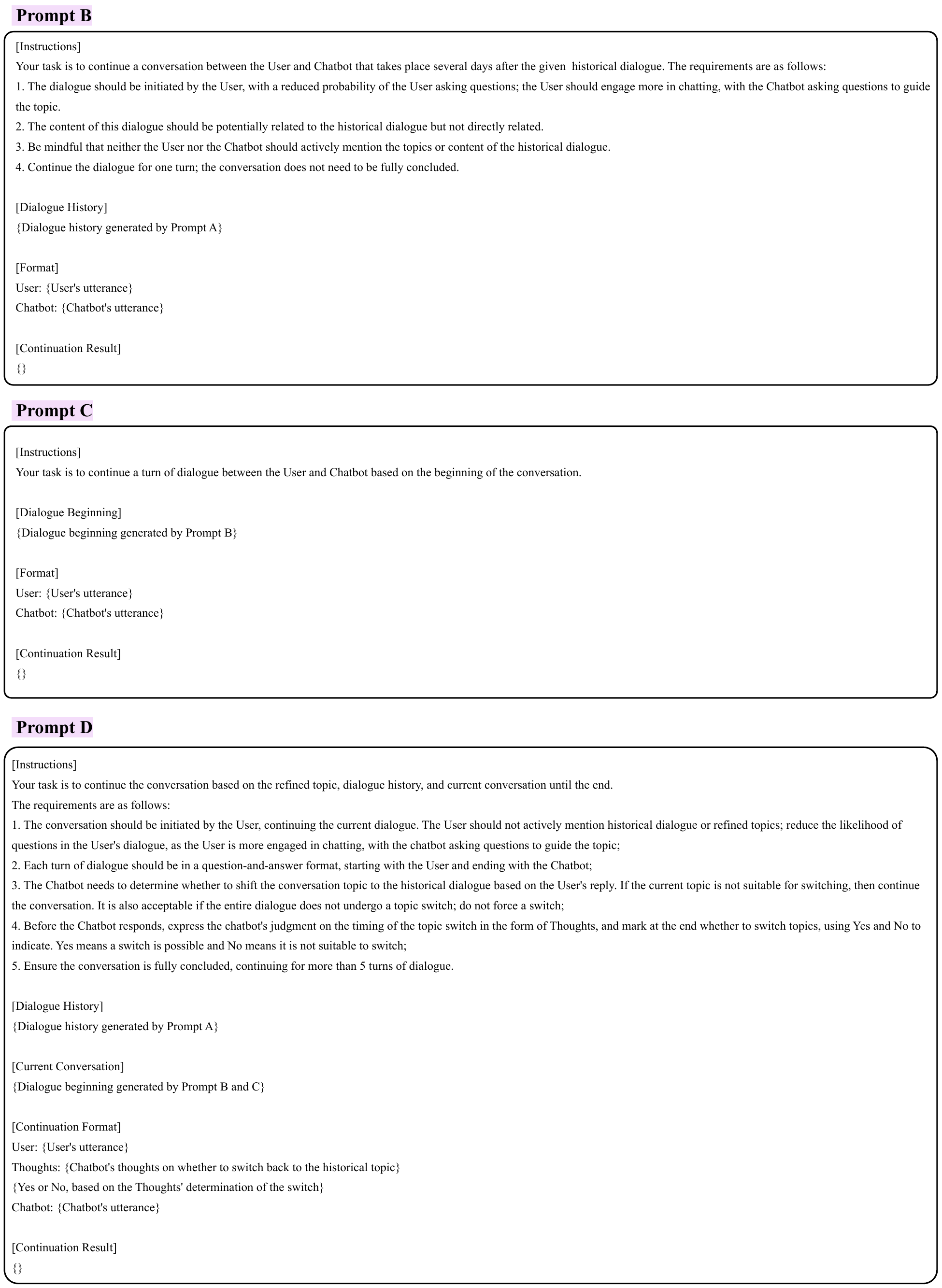}
    \caption{
     English version of prompt for data construction in Figure \ref{prompt1}.
    }
    \label{prompt2}
    \end{figure*}

\begin{table}[htbp]
\centering
\renewcommand{\arraystretch}{1.1}
\resizebox{\linewidth}{!}{
\begin{tabular}{p{2.3cm}ccccc}
\toprule
 & \textbf{ROUGE-1} & \textbf{ROUGE-2} & \textbf{ROUGE-L} & \textbf{BLEU$^{1-4}$} & \textbf{METEOR} \\ \hline
Qwen2.5 & 0.522 & 0.333 & 0.467 & 0.197 & 0.414 \\
Qwen2.5-Sum & \textbf{0.773} & \textbf{0.646} & \textbf{0.746} & \textbf{0.536} & \textbf{0.755} \\ \hline
\toprule
\end{tabular}} 
\caption{Comparison results of the Qwen model with and without fine-tuning on our ChMapData-Sum dataset.
}
\label{tab:result_sum}
\end{table}

\section{ChMapData-test Construction}
\label{sec:construction}
The test set construction process is consistent with Section \ref{data_construction}.
Initially, we generated 400 dialogues from 11 topics, ensuring the same ratio of Memorable and General data in the training set.
Subsequently, 150 dialogues under the Memorable Subject were extended by two turns, serving as the beginning of the current context. 
Consistent with the trainset construction, we utilized the corresponding topics generated in the first step and chose other 29 historical topics at random as candidates to rank against the current context. 
Both the corresponding topic and the top-ranked topic were utilized as ground truth, as they each acted as positive examples in the training set. Additionally, 2 to 11 dialogues were randomly chosen from topics ranked below the corresponding topic (i.e., negative examples) to serve as dialogue history. 
This process resulted in 112 entries\footnote{Excluding 22 entries with fewer than two topics ranked below the corresponding and 16 items modified after ranking.}, out of which 100 were finalized for the test set, each comprising dialogue history, current dialogue, and a ground truth topic.

\section{Comparison Result for Summarization Model}
\label{sec:comp_result}

We compared the performance of our Qwen2.5-Sum, a fine-tuned version of Qwen2.5 on our \textit{ChMapData-Sum}, against the original Qwen2.5 model. We reported standard automated metrics including BLEU \citep{papineni2002bleu}, ROUGE \citep{lin-2004-rouge}, and METEOR \citep{banerjee2005meteor}. Specifically, we provided the full BLEU score, which accounts for overlap across 1-4 grams, rather than just BLEU-4. 
The results, as shown in Table \ref{tab:result_sum}, indicate that Qwen2.5-Sum significantly outperforms the original Qwen2.5 model, demonstrating the effectiveness of our summarization module.

\section{Implementation Details}
\label{sec:imple}

For experiments on Topic Summarization, Topic Retrieval, and Proactive Topic-shifting Detection and Generation, we utilized the AdamW optimizer \citep{loshchilov2019decoupled}. The training setup included a cosine learning rate schedule starting at 2e-5, a weight decay of 0.1, a batch size of 64, a 5\% warm-up period, and a maximum input length of 2048 tokens. We fine-tuned all the models for 2 epochs.

\section{Prompt for Proactive Dialogue Models}
\label{sec:appendix_model_prompt}
\begin{figure*}
    \centering
    \includegraphics[width=\linewidth]{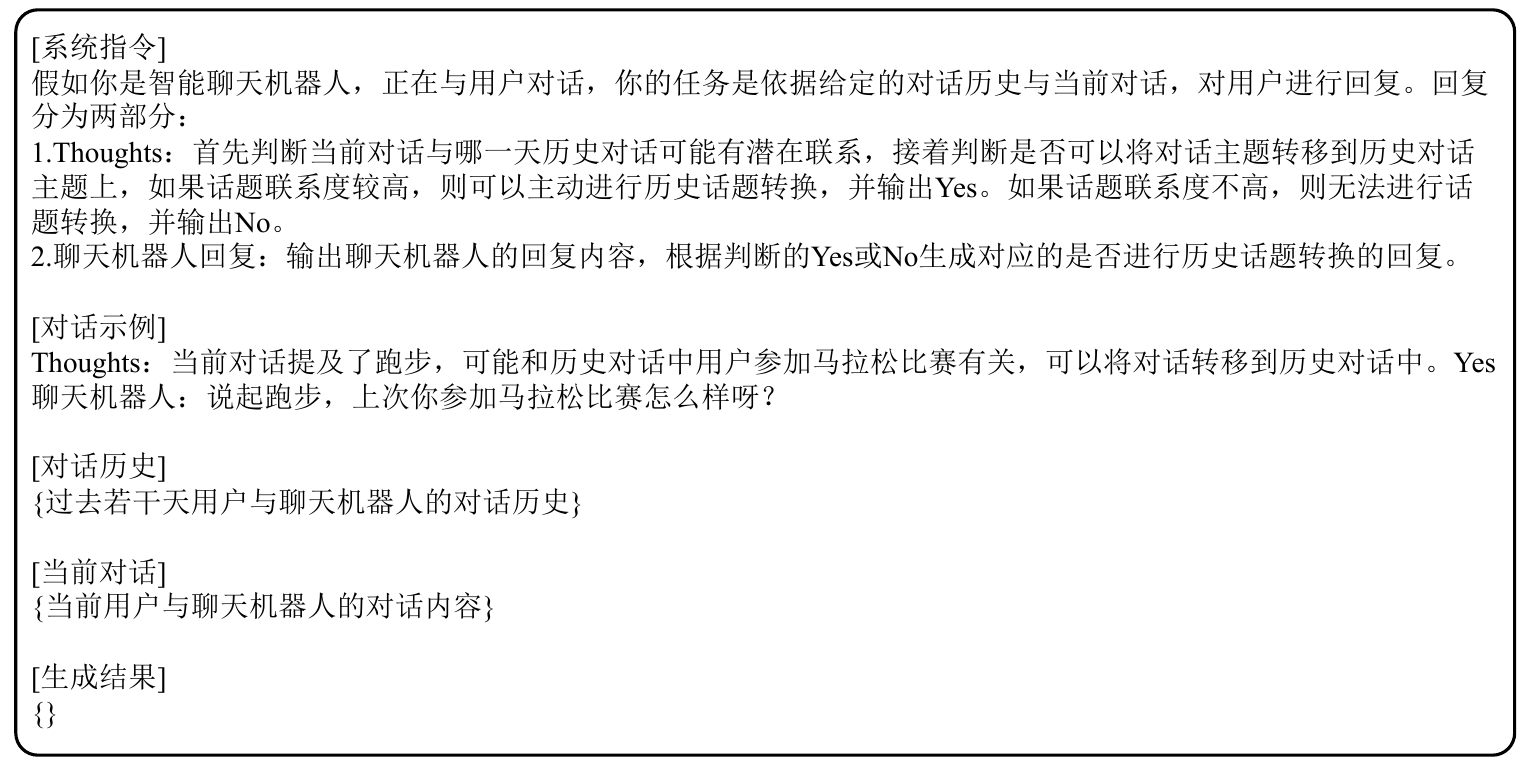}  
    \includegraphics[width=\linewidth]{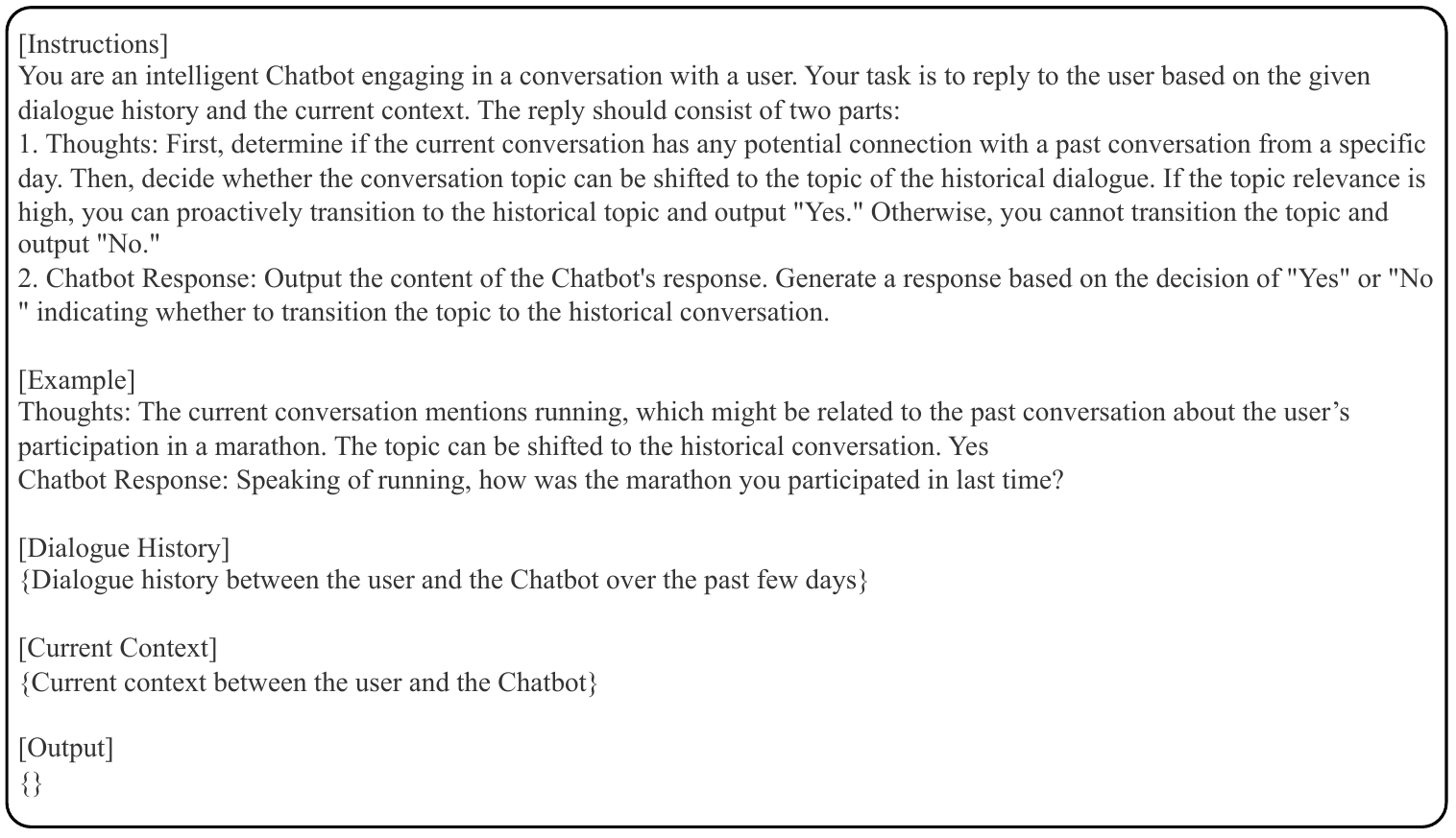}
    \caption{
    The prompt template instructs Qwen-2.5 and GPT-4 to act as the en-to-end memory-aware proactive dialogue system. Upper is the original content input into the model, followed by its corresponding English version.}  
    \label{model_prompt}
    \end{figure*}

\begin{figure*}
    \centering
    \includegraphics[width=\linewidth]{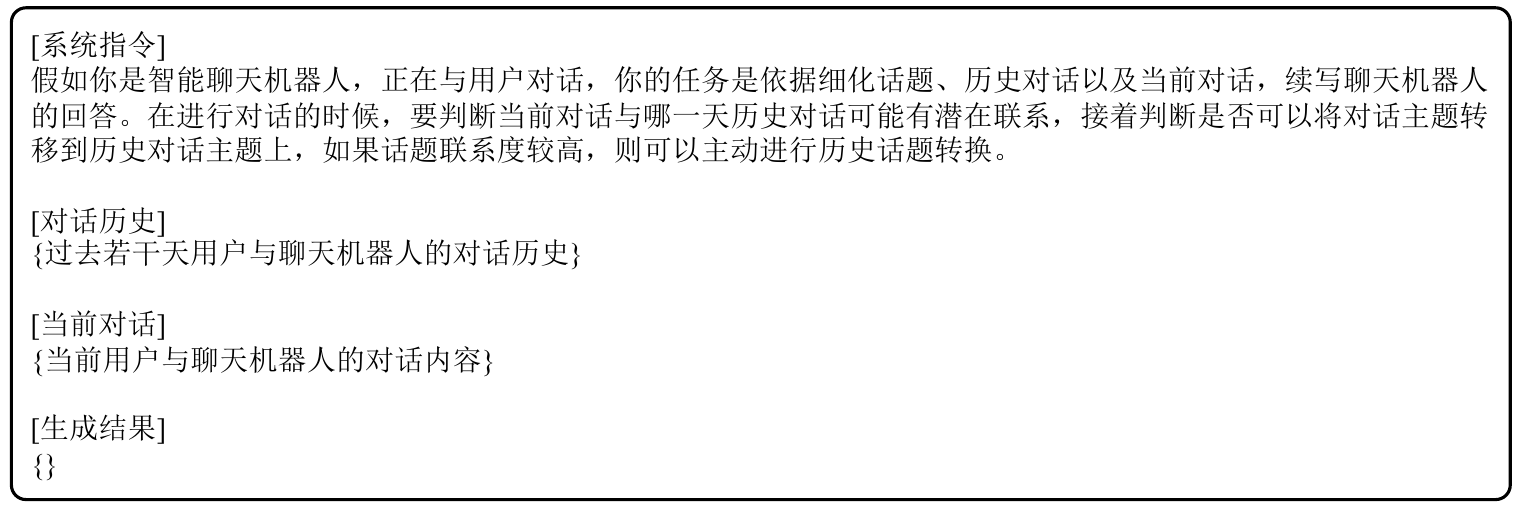}
    \includegraphics[width=\linewidth]{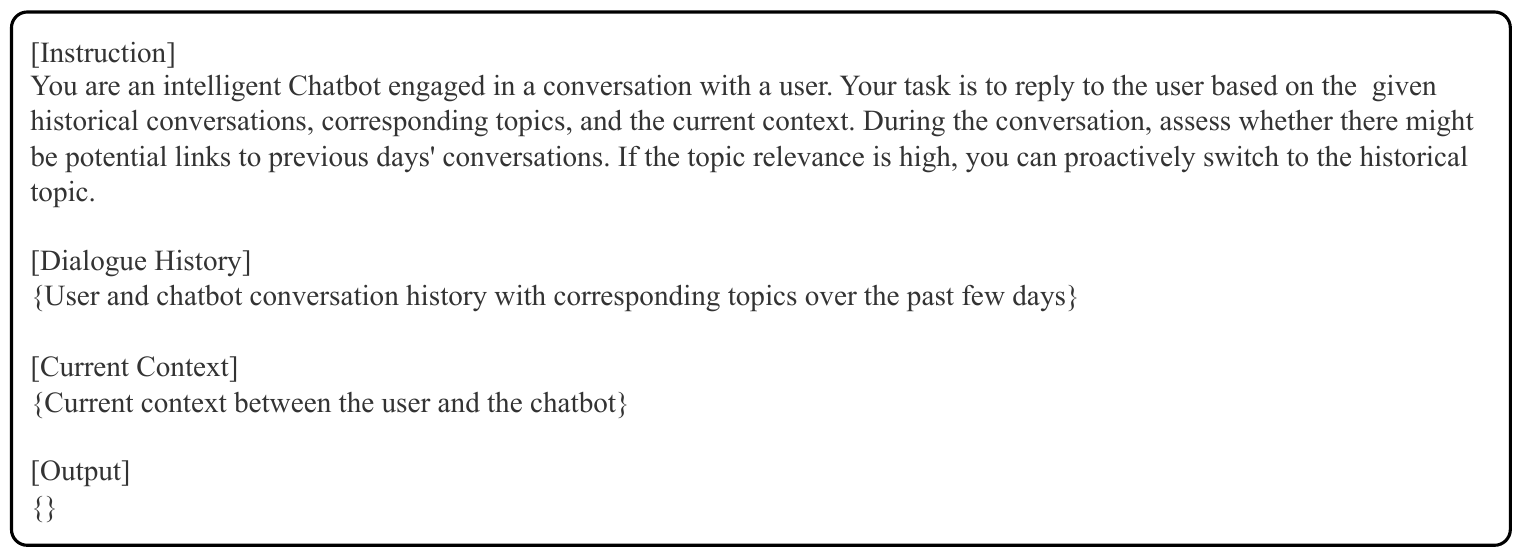}
    \caption{
    The prompt template for the original Qwen2.5 is used to generate a proactive dialogue response along with its English version.
    }    
    \label{model_prompt_1}
    \end{figure*}

In Figure \ref{model_prompt}, we present the full prompt templates for the two models, Qwen2.5 and GPT-4, which function as memory-aware proactive dialogue systems in Section \ref{subsec: approaches}. Additionally, Figure \ref{model_prompt_1} illustrates the prompt used within the BGE w/ Qwen framework for the original Qwen2.5 model, guiding it to generate proactive dialogue responses.

\section{Details of User-role dialogue model}
\label{sec: user}
We additionally trained a dialogue model to simulate user interactions during model testing. This approach helps to avoid the subjective factors that annotators might introduce during conversations, which could affect the guidance of active topics.

The parameters of the user-role dialogue model are consistent with those in Appendix \ref{sec:imple}. We utilize Qwen2.5-7B as the base model and the data used to train the user model consists of 4,000 dialogue histories generated in Section \ref{data_construction} Step 1. We performed additional processing on the data by converting the training target to the query rather than the response. Moreover, to ensure that the user model does not prematurely end the conversation, we removed the last round from the training data, as this turn typically signifies the end of the conversation.

\section{Human Annotation Details}
\label{sec:appendix_guideline}
Table \ref{tab_guideline} presents our full annotation guidelines used for the human annotation process in this work.

We recruited six college students who are native Chinese speakers, including four females and two males, with an average age of around 24. Initially, they were provided with an annotation guideline. Each evaluator underwent a training process to enhance their understanding of the annotation procedure. Before starting the annotation, we designed a qualification test consisting of 10 dialogues; only those who passed the test were deemed qualified and allowed to proceed with the annotation
To ensure the quality of the annotations, we divided the dataset into batches and assigned a specific number of daily tasks to each annotator. Upon receiving the daily annotations, we reviewed the results and required annotators to reannotate the batch of data assigned for that day if there was low agreement (less than 0.6).

In the annotation interface, the dialogue history, summarized topic, and current context were presented on the left side, while the dialogues generated by each model were randomly displayed on the right to prevent bias. 
Annotators first read each chatbot's utterance and then assigned scores for "Engagingness" and "Overall quality." After completing the entire session, they assessed "Achievement" and "\#Turn." The score range for the first three evaluation criteria was \{0,1,2\}, while the range for "\#Turn" was 0-10.

Each sample was annotated by two distinct annotators, and a third annotator made the final decision in case of disagreement. We utilized Cohen's kappa \citep{Cohen1960ACO} to measure inter-annotator agreement. The annotation process lasted approximately two weeks, culminating in a substantial inter-annotator agreement with Cohen’s kappa of $\kappa$=0.70, as shown in Table \ref{tab:overall_eval}.

\section{Integration Testing}
\label{sec:integration}

\begin{table}[h]
\centering
\begin{tabular}{llc}
\toprule
\textbf{Model} & \textbf{CPS} & \textbf{Shift-Ratio} \\ \hline
Original & 22.8 & - \\
Original w/ PDia & 25.0 & 20.8\% \\
Original w/ MapDia & \textbf{27.9} & 12.2\% \\
\toprule
\end{tabular}
\caption{Results of integrated testing, showing the conversation-turns-per-session (CPS) and the triggered ratio of topic shifts per session in a real-world dialogue system. The p-value for the CPS statistic is 0.0074.
}
\label{tab10}
\end{table}

Given the novel method proposed in this paper, which can detect conversation trajectories and initiate proactive topics based on dialogue history, it is highly adaptable for integration with any existing dialogue system. To assess its effectiveness, we conducted an online A/B test by incorporating it into our role-playing dialogue system. The proactive model is trained with combined data of role-playing conversational dataset and ChMapData-Mem to keep the role-playing ability. The model determines when to shift the topic and generates responses for those turns, while the original dialogue system handles other responses. 

Additionally, we conducted another integration with trained a proactive responding model only referring to the dialogue context, noted as Proactive Dialogue (PDia). PDia shares the pre-trained base and parameter scale as our proposed model. This model also employs targeted data construction followed by fine-tuning to learn proactive dialogue capabilities. Besides, different from traditional proactive dialogue methods by performing topic planning in advance, we utilize the LLM to dynamically make decisions during the dialogue process.

Given that, users were randomly assigned to one of three groups: one interacting with the original dialogue system, one with the system enhanced by our proposed method, and one with the context-based proactive model. All users were blinded to the system details. Due to commercial constraints, we utilized a closed-source 7B pre-trained model for retraining the proactive dialogue models.

\begin{table}[h]
\centering
\small
\renewcommand{\arraystretch}{1.1}
\resizebox{\linewidth}{!}{
\begin{tabular}{lcccccr}
\toprule
 & \multicolumn{2}{c}{\textbf{Engagement}} &  & \multicolumn{2}{c}{\textbf{Achievement}} & \multicolumn{1}{c}{} \\
\multirow{-2}{*}{\textbf{Models}} & \textit{Utts.-level} & \textit{Sess.-level} & \multirow{-2}{*}{\textbf{\begin{tabular}[c]{@{}c@{}}Overall \\ Quality\end{tabular}}} & Mentioning & Shifting & \multicolumn{1}{c}{\multirow{-2}{*}{\textbf{Avg.}}} \\ \hline
\multicolumn{7}{c}{\cellcolor[HTML]{EFEFEF}\#Turn = 1} \\ \hline
Ours per Sess. & 0.38 & 0.71 & 1.11 & 0.11 & 0.89 & 1.19 \\
Ours per Utts. & 0.31 & 0.58 & 1.05 & 0.16 & 0.84 & 1.07 \\
GPT-4 & 0.43 & 0.46 & 0.89 & 0.09 & 0.91 & 1.02 \\ \hline
\multicolumn{7}{c}{\cellcolor[HTML]{EFEFEF}\#Turn = 2} \\ \hline
Ours per Sess. & 0.31 & 0.50 & 1.10 & 0.21 & 0.79 & 1.07 \\
Ours per Utts. & 0.42 & 0.55 & 1.07 & 0.18 & 0.82 & 1.10 \\
GPT-4 & \textbf{0.55} & 0.53 & 1.12 & 0.21 & 0.79 & 1.13 \\ \hline
\multicolumn{7}{c}{\cellcolor[HTML]{EFEFEF}\#Turn = 3} \\ \hline
Ours per Sess. & 0.32 & 0.58 & 1.18 & \textbf{0.33} & 0.67 & 1.05 \\
Ours per Utts. & 0.35 & 0.64 & 1.13 & 0.21 & 0.79 & 1.09 \\
GPT-4 & 0.43 & 0.64 & 0.89 & 0.21 & 0.79 & 1.08 \\ \hline
\multicolumn{7}{c}{\cellcolor[HTML]{EFEFEF}\#Turn = 4} \\ \hline
Ours per Sess. & 0.31 & 0.33 & 1.28 & 0.22 & 0.78 & 1.04 \\
Ours per Utts. & 0.38 & \textbf{0.80} & \textbf{1.40} & 0.00 & \textbf{1.00} & \textbf{1.21} \\
GPT-4 & 0.35 & 0.43 & 0.91 & 0.29 & 0.71 & 0.98 \\ 
\toprule
\end{tabular}
}
\caption{Evaluation results for each turn number at which the model shifts topics.
}
\label{tab7}
\end{table}

\begin{table}[h]
\centering
\renewcommand{\arraystretch}{1.1}
\resizebox{\linewidth}{!}{
\begin{tabular}{llc}
\toprule
\textbf{Model} & \textbf{\begin{tabular}[c]{@{}l@{}}Retrieval\\ Method\end{tabular}} & \textbf{Mentioning} \\ \hline
\multirow{2}{*}{QSum w/ BGE w/ Qwen} & per Sess. & 0.07 \\
 & per Utt. & 0.08 \\ \hline
\multirow{2}{*}{QSum w/ QRet w/ Qwen} & per Sess. & 0.08 \\
 & per Utt. & 0.15 \\ \hline
\multirow{2}{*}{Ours} & { per Sess.} & 0.18 \\
 & per Utt. & 0.14 \\ \hline
GPT-4 & \textbf{-} & 0.19 \\ 
\toprule
\end{tabular}
}
\caption{Probability of each model mentioning historical topics, calculated as the proportion of label 1 in the Achievement criteria.
}
\label{tab8}
\end{table}

The test spanned a duration of two weeks and involved real conversations from over 100,000 users. Table~\ref{tab10} presents the conversation-turns-per-session (CPS), defined as the average number of conversation turns between the dialogue system and the user within a session. The introduction of both proactive topic capabilities significantly enhanced CPS. Specifically, the MapDia model increased the average CPS from 22.8 to 31.3, which is notably higher than that of PDia, indicating that users are more engaged with previously discussed topics when they are properly introduced.

Additionally, it should be noted that the proportion of topic transitions is significantly lower than reported in Table~\ref{tab:overall_eval}. 
This discrepancy is primarily attributed to the fact that only a small portion of real user dialogues can effectively integrate previously discussed content, and not all conversations require the initiation of proactive topics. Even the PDia model, which incorporates dialogue context, successfully transitions topics in only 20.8\% of sessions.

\section{Analysis of \#Turn and Mentioning Metrics}
\label{sec:turn}

Here, we present the evaluation metrics for \#Turns set at 1, 2, 3, and 4 in Table \ref{tab7}. Our model demonstrated the best performance in turn-level retrieval when transitioning topics in the fourth turn. It is observed that fewer \#Turns may result in lower overall quality and lower average scores. There is no distinct proportional or inverse correlation between the number of \#Turns and the model's overall performance. 

Table \ref{tab8} additionally shows the probability of the model mentioning historical topics without transitioning, which indicates that GPT-4 is more inclined to mention historical topics, which deviates from our task definition.

\begin{table*}[t]
\centering
\setlength{\extrarowheight}{0pt}
\addtolength{\extrarowheight}{\aboverulesep}
\addtolength{\extrarowheight}{\belowrulesep}
\setlength{\aboverulesep}{0pt}
\setlength{\belowrulesep}{0pt}
\resizebox{\linewidth}{!}{
\begin{tabular}{|p{18cm}|} 
\hline
\multicolumn{1}{|c|}{\cellcolor[rgb]{0.9,0.9,0.9}{\large\textbf{Human Evaluation Guideline}}}                                                                                                                                         \\ 
\hline
\textbf{\large\textcolor{red}{Task Overview}}                                                                                                                                                                               \\
Thank you for participating in this task! Open-domain dialogue systems are expected to possess the capability to proactively shift conversational topics when necessary. When a chat agent exhausts its conversational material or the current discussion becomes monotonous, topic shifting is a common strategy to maintain the flow of conversation. Furthermore, when the new topic is derived from historical conversations rather than arbitrary subjects, it enhances user engagement and fosters long-term relationships between the chatbot and the user.
To achieve this objective, we have developed a Memory-aware Proactive Dialogue system. Below, we provide several days' worth of historical dialogues, along with responses generated by our model and some baseline models. Your task is to evaluate these responses based on the four defined aspects.
\\                                                                                                                                                                                                                                       \\ 
\hline
\textbf{\large\textcolor{red}{Evaluation Aspects}}                                                                                                                                                                                                                                                             \\

\begin{tabular}[c]{@{}p{18cm}@{}}\large\textbf{Utterance-level} \\\begin{tabular}{@{\labelitemi\hspace{\dimexpr\labelsep+0.5\tabcolsep}}p{16.5cm}@{}}	\textbf{Engagingness:}  An utterance-level metric measuring how well the chatbot captures and responds to the user's personal attributes, preferences, and interests, encouraging ongoing participation and long-term connections.~ ~ ~~\end{tabular}\end{tabular}                                                                                                                                                                                                                            \\
\begin{tabular}{@{\labelitemi\hspace{\dimexpr\labelsep+0.5\tabcolsep}}p{17.5cm}@{}}	\textbf{Overall Quality:}An utterance-level metric assessing the effectiveness of the response, focusing on smooth topic transitions, coherence, and consistency with the given context.~ ~ ~~\end{tabular}
\\

\begin{tabular}[c]{@{}p{18cm}@{}}\large\textbf{Session-level} \\\begin{tabular}{@{\labelitemi\hspace{\dimexpr\labelsep+0.5\tabcolsep}}p{17.5cm}@{}}	\textbf{Achievement:} A session-level metric determining whether the chatbot has successfully transitioned the conversation to a relevant historical topic. ``1'' indicates the topic is merely mentioned, and ``2'' signifies a successful transition. ~ ~ ~~\end{tabular}\end{tabular}                                                                                                                                                                                                                            \\
\begin{tabular}{@{\labelitemi\hspace{\dimexpr\labelsep+0.5\tabcolsep}}p{17.5cm}@{}}	\textbf{Turn:} A session-level metric represents the turn number when mention or transition occurred.~ ~ ~~\end{tabular}
\\                                                                                
                                                                                                                                                       \\ 
\hline
\textbf{\large\textcolor{red}{Annotation Procedure}}                              \\
\begin{tabular}{@{\hspace{\dimexpr\labelsep+0.5\tabcolsep}}p{17.5cm}@{}}	\textbf{1.	Dialogue History Familiarization:} Begin by thoroughly reading and familiarizing yourself with the provided historical dialogues, typically spanning 8-10 days.~ ~ ~~\end{tabular}
\\
\begin{tabular}{@{\hspace{\dimexpr\labelsep+0.5\tabcolsep}}p{17.5cm}@{}}	\textbf{2.	Current Context Review:} Carefully read the initial context of the current dialogue, which includes two beginning turns and a user utterance.~ ~ ~~\end{tabular}
\\
\begin{tabular}{@{\hspace{\dimexpr\labelsep+0.5\tabcolsep}}p{17.5cm}@{}}	\textbf{3.	Utterance Scoring:} Score each response utterance generated by the model on a scale of [0, 1, 2] based on the aspects of Engagingness and Overall Quality. A higher score indicates better performance.~ ~ ~~\end{tabular}
\\
\begin{tabular}{@{\hspace{\dimexpr\labelsep+0.5\tabcolsep}}p{17.5cm}@{}}	\textbf{4.	Session Scoring:} Once the model completes the dialogue continuation, determine whether the entire session achieved a topic shift. Here, 0 indicates no topic shift or mentioning, 1 indicates a mention of a historical topic without shifting, and 2 indicates a complete topic shift. Additionally, note the turn number at which the shift was accomplished. ~ ~ ~~\end{tabular}\\

                                                                                                                                                       \\ 
\hline
\textbf{\large\textcolor{red}{Emphasis and Caution}}                              \\
\begin{tabular}{@{\labelitemi\hspace{\dimexpr\labelsep+0.5\tabcolsep}}p{17.5cm}@{}}The order of the model-generated responses is randomized to avoid bias.~ ~ ~~\end{tabular}
\\
\begin{tabular}{@{\labelitemi\hspace{\dimexpr\labelsep+0.5\tabcolsep}}p{17.5cm}@{}}	It is possible for the model to perform multiple topic shifts within a single session. This strength can be reflected by assigning a score of 2 for Engagingness or Overall Quality at each turn where a topic shift occurs.~ ~ ~~\end{tabular}
\\
\begin{tabular}{@{\labelitemi\hspace{\dimexpr\labelsep+0.5\tabcolsep}}p{17.5cm}@{}}When the topic shift is not natural or smooth, the Overall Quality score should be appropriately reduced, even if the shift was achieved.~ ~ ~~\end{tabular}
\\
\begin{tabular}{@{\labelitemi\hspace{\dimexpr\labelsep+0.5\tabcolsep}}p{17.5cm}@{}}	A number of words and phrases are often used as indicators for topic shifts, including but not limited to: "but," "speaking of," "talking about," "anyway," "by the way," "that reminds me," "before I forget," "I want to mention," "let’s talk about," "we need to discuss," "funny you should mention that", etc. ~ ~ ~~\end{tabular}\\

\\
\hline
\end{tabular}
}
\caption{The full annotation guideline for human evaluation.
}
\label{tab_guideline}
\end{table*}

\end{document}